\documentclass{article}
\usepackage{arxiv}
\usepackage[utf8]{inputenc} % allow utf-8 input
\usepackage[T1]{fontenc}    % use 8-bit T1 fonts
\usepackage{hyperref}       % hyperlinks
\usepackage{url}            % simple URL typesetting
\usepackage{booktabs}       % professional-quality tables
\usepackage{amsfonts}       % blackboard math symbols
\usepackage{nicefrac}       % compact symbols for 1/2, etc.
\usepackage{microtype}      % microtypography
\usepackage{lipsum}
\usepackage{graphicx}
\graphicspath{ {./images/} }
\usepackage{subfigure}
\usepackage{color, colortbl}
\usepackage{times,amsmath,epsfig,bm,float,epstopdf,graphicx,graphics,amsthm}
\usepackage{balance,multicol,multirow}
\usepackage{color}
\usepackage{verbatim}
\usepackage{algorithm, algorithmic, enumerate}
\usepackage{amsfonts}
\usepackage{booktabs}
\usepackage[inline]{enumitem}
\usepackage{blindtext}
\usepackage{gensymb}
\usepackage{chemformula}
\usepackage{siunitx}
\usepackage{colortbl}
\usepackage{amsmath}
\usepackage{cite}
\usepackage{siunitx}
\usepackage{gensymb}
\usepackage{float}
\usepackage{enumitem}
\usepackage{amsmath}
\usepackage{multirow}

\usepackage{graphics}
\usepackage{csquotes}
\usepackage{makecell}

\title{Creation of AI-driven Smart Spaces for Enhanced Indoor Environments -- A Survey}

\author{
 Aygün Varol$^*$\\
  Department of Computing Sciences\\
  Tampere University\\
  Tampere, Finland\\
  \texttt{aygun.varol@tuni.fi} \\
    \And
 Naser Hossein Motlagh$^*$\\
  Department of Computer Science\\
  University of Helsinki, Finland\\
  Helsinki, Finland \\
  \texttt{naser.motlagh@helsinki.fi} \\
\And
  Mirka Leino\\
  Faculty of Technology\\
  Satakunta University of Applied Sciences\\
  Pori, Finland\\
  \texttt{mirka.leino@samk.fi} \\
  \And
  Sasu Tarkoma \\
  Department of Computer Science\\
  University of Helsinki\\
  Helsinki, Finland  \\
  \texttt{sasu.tarkoma@helsinki.fi} \\
    \And
  Johanna Virkki\\
  Department of Computing Sciences\\
  Tampere University\\
  Tampere, Finland \\
  \texttt{johanna.virkki@tuni.fi} \\
}

\begin{document}
\maketitle

\let\thefootnote\relax
\footnotetext{$^*$Both authors contributed equally to this research.} %%%%%%%%%%

\begin{abstract}
Smart spaces are ubiquitous computing environments that integrate diverse sensing and communication technologies to enhance space functionality, optimize energy utilization, and improve user comfort and well-being. The integration of emerging AI methodologies into these environments facilitates the formation of AI-driven smart spaces, which further enhance functionalities of the spaces by enabling advanced applications such as personalized comfort settings, interactive living spaces, and automatization of the space systems, all resulting in enhanced indoor experiences of the users. In this paper, we present a systematic survey of existing research on the foundational components of AI-driven smart spaces, including sensor technologies, data communication protocols, sensor network management and maintenance strategies, as well as the data collection, processing and analytics. Given the pivotal role of AI in establishing AI-powered smart spaces, we explore the opportunities and challenges associated with traditional machine learning (ML) approaches, such as deep learning (DL), and emerging methodologies including large language models (LLMs). Finally, we provide key insights necessary for the development of AI-driven smart spaces, propose future research directions, and sheds light on the path forward.
\end{abstract}


\section{Introduction}
%=====================
\label{sec:Introduction}

Smart spaces are defined as ubiquitous computing environments that enhance indoor settings by seamlessly integrating sensors, actuators, and communication technologies into the physical environment to deliver intelligent services~\cite{aziz2016requirement}. These systems leverage sensing mechanisms, communication protocols, and pervasive computing approaches to automate various indoor functions, including lighting and Heating, Ventilation, and Air Conditioning (HVAC) systems. Based on the literature, such environments also support real-time data processing and analytics to create intelligent, adaptive living spaces~\cite{motlagh2023digital}.

While existing smart spaces provide significant automation benefits, their designs often prioritize sensor-driven responses over user-centric needs. These systems typically rely on rule-based automation and manual configuration—such as turning lighting systems on or off based on predefined thresholds—without adapting dynamically to user preferences or behaviors~\cite{goyal2021internet}. Consequently, they fall short in utilizing the extensive data generated by the diverse array of sensors deployed in these environments. The lack of robust mechanisms for interpreting data and generating actionable insights limits their capacity to transcend basic automated actions and evolve into truly adaptive systems.

The integration of Artificial Intelligence (AI) into smart spaces holds the potential to revolutionize these environments, enabling them to transition into fully intelligent ecosystems. AI-driven smart spaces can support real-time decision-making, predict and respond to user needs, and enhance personalization. By leveraging AI, these spaces can achieve higher efficiency, functionality, and user-centric adaptability, leading to improved safety, well-being, and comfort for residents.

\begin{figure}%[t]
    \centering
   \includegraphics[width=.8\columnwidth]{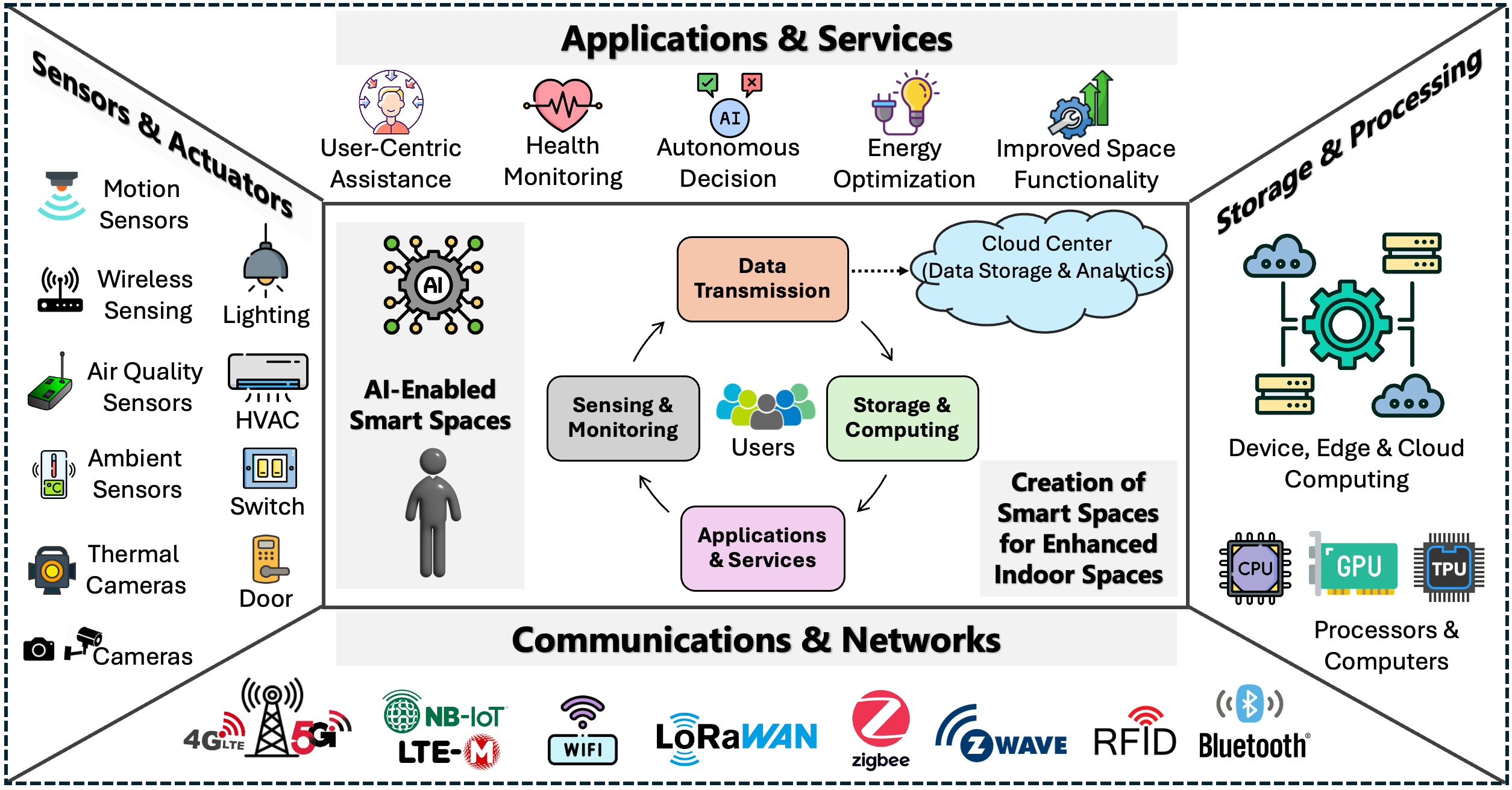}
    \caption{The necessary components needed for creation of AI-driven smart spaces.}
    \label{fig:overview}
    \vspace{-.5cm}
\end{figure}

AI-driven smart spaces enable diverse applications, including activity monitoring, anomaly detection, energy optimization~\cite{sleem2023survey, almusaed2023enhancing, nwakanma2023review, lee2024use, sutjarittham2019experiences}, occupancy detection~\cite{abade2018non, han2018smart, saleem2023smart, motlagh2021monitoring}, user behavior prediction~\cite{almeida2022comparative, zhang2023multisense}, interactive environment creation~\cite{khelifi2024mixed}, personalized recommendation systems~\cite{wu2024survey}, and advanced care solutions~\cite{jovan2023multimodal}. They build upon existing smart space technologies and integrate emerging methodologies, such as Internet of Things (IoT) frameworks, machine learning (ML), deep learning (DL), transformer networks, and large language models (LLMs). These advancements enable continuous learning from user behaviors to optimize operations, fostering functional, interactive, healthy, and sustainable indoor environments.

Despite their promise, developing AI-driven smart spaces poses several challenges alongside significant opportunities. This article provides a comprehensive review of the components and technologies —such as sensor systems and communication protocols with an emphasize on emerging AI methodologies such as transformer networks and LLMs— essential for building AI-driven smart spaces. We synthesize current research, analyze existing methodologies and solutions, and identify emerging trends alongside the challenges and limitations of each technology. To advance the development of AI-enabled smart spaces, we also propose future research directions and present a roadmap for the path forward.

\section{Scope of Survey}
%=====================
\label{sec:Scope of Survey}

\subsection{Related Surveys}
Due to the growing popularity and importance of smart spaces, numerous studies have been conducted in this domain. Table~\ref{tab:existing-surveys} provides a summary of existing works that intersect with our study. 

\begin{table}%[ht]
    \centering
    \caption{Existing related surveys in the literature.}
    \resizebox{.8\columnwidth}{!}{
    \begin{tabular}{l l}
         \toprule
         \textbf{Scope} & \textbf{Survey}\\
         \midrule 
         \multirow{1}{*}{Sensing applications}
         & Artificial Intelligence~\cite{siam2024artificial, lu2024application, alsafery2023sensing}\\
         & \\
         \hline 
         \multirow{1}{*}{Smart lighting}
         & Conventional ML and DL models~\cite{putrada2022machine}\\
         & \\
         \hline 
         \multirow{2}{*}{Human activity recognition}
         & Conventional ML models~\cite{diraco2023human, gu2021survey}\\
         & Resource-constrained IoT environments~\cite{diraco2023review, babangida2022internet, bouchabou2021survey}\\
                  & \\
         \hline
         \multirow{1}{*}{Energy optimization in buildings}
         & Deep Reinforcement Learning models~\cite{yu2021review, zhang2022building, nagy2023ten, lissa2021deep, shaqour2022systematic} \\
                  & \\
         \hline
         \multirow{1}{*}{Air quality and energy management}
         & Deep Reinforcement Learning models~\cite{ogundiran2024systematic, tien2022machine, merabet2021intelligent}\\
                  & \\
         \bottomrule
    \end{tabular}
    }
    \label{tab:existing-surveys}
\end{table}

State-of-the-art research often emphasizes survey studies focusing on practical machine learning (ML) and deep learning (DL) applications, addressing the need for diverse sensing systems to enable smart space applications~\cite{lu2024application}. 
Some studies specifically survey ML techniques for smart lighting applications~\cite{putrada2022machine} or introduce sensor technologies and deployment strategies leveraging AI methodologies~\cite{siam2024artificial, alsafery2023sensing}. Other research concentrates on human activity recognition (HAR) within smart spaces, exploring public datasets, sensor requirements, and traditional ML models~\cite{diraco2023human, gu2021survey}. Surveys also explore HAR applications in resource-constrained smart living environments, focusing on sensing technologies and sensor communication frameworks~\cite{diraco2023review,babangida2022internet, bouchabou2021survey}. 
In addition, several studies examine the application of AI methodologies such as deep reinforcement learning (DRL) for energy management in smart buildings, addressing challenges such as energy efficiency, carbon emission reduction, and user comfort~\cite{yu2021review, zhang2022building, nagy2023ten, lissa2021deep, shaqour2022systematic}. Other research investigates indoor air quality, correlating it with energy consumption and demonstrating how AI enhances air quality while optimizing energy utilization in smart spaces~\cite{ogundiran2024systematic, tien2022machine, merabet2021intelligent}.

While the existing literature offers valuable insights, it has notable limitations. First, most surveys narrow their scope to specific applications, such as human activity recognition or energy optimization. Second, they often overlook essential components such as communication technologies, protocols, and data collection and processing strategies, which are critical for developing AI-powered smart spaces. Third, the intricacies of advanced AI models, such as Transformer networks, are frequently neglected, despite their pivotal role in transforming smart spaces into fully intelligent and interactive spaces.

In contrast, our work takes a broader perspective, encompassing diverse applications of smart spaces. We introduce the foundational components required to build AI-driven smart environments and highlight the significance of both conventional ML approaches and emerging technologies, such as Transformer networks and large language models (LLMs). By providing a comprehensive overview, our study underscores the deployment of cutting-edge technologies in creating intelligent, interactive, and efficient smart spaces.

\subsection{Selection of Articles}

To provide a comprehensive overview of AI-driven smart spaces, we identified the essential components required to create such spaces, explored their applications in indoor environments, and examined the role of AI techniques in empowering these systems. Our methodology involved conducting database queries to select relevant articles for inclusion in our survey. 
We queried databases including Google Scholar, IEEE Xplore, ACM Digital Library, and ScienceDirect using combinations of the following keywords: \textit{artificial intelligence, machine learning, deep learning, transformer networks, generative pre-trained transformer networks, large language models, long short-term memory, recurrent neural networks, natural language processing, smart indoor environment, smart space, smart home, smart building, smart house, and smart office}. 
For instance, we utilized search combinations such as \enquote{LLMs in smart homes} and \enquote{Utilization of transformer networks in smart indoor spaces} to retrieve relevant articles. From the search results, we prioritized articles that were highly cited, frequently referenced, or had themselves cited other relevant works. Each selected article was thoroughly examined and categorized based on its relevance to the topic and scope of our survey. Ultimately, these labeled and relevant articles formed the foundation of our comprehensive review of AI-driven smart spaces.

\subsection{Survey Structure}

\begin{figure*}[ht]
  \centering
    \includegraphics[width=0.95\linewidth]{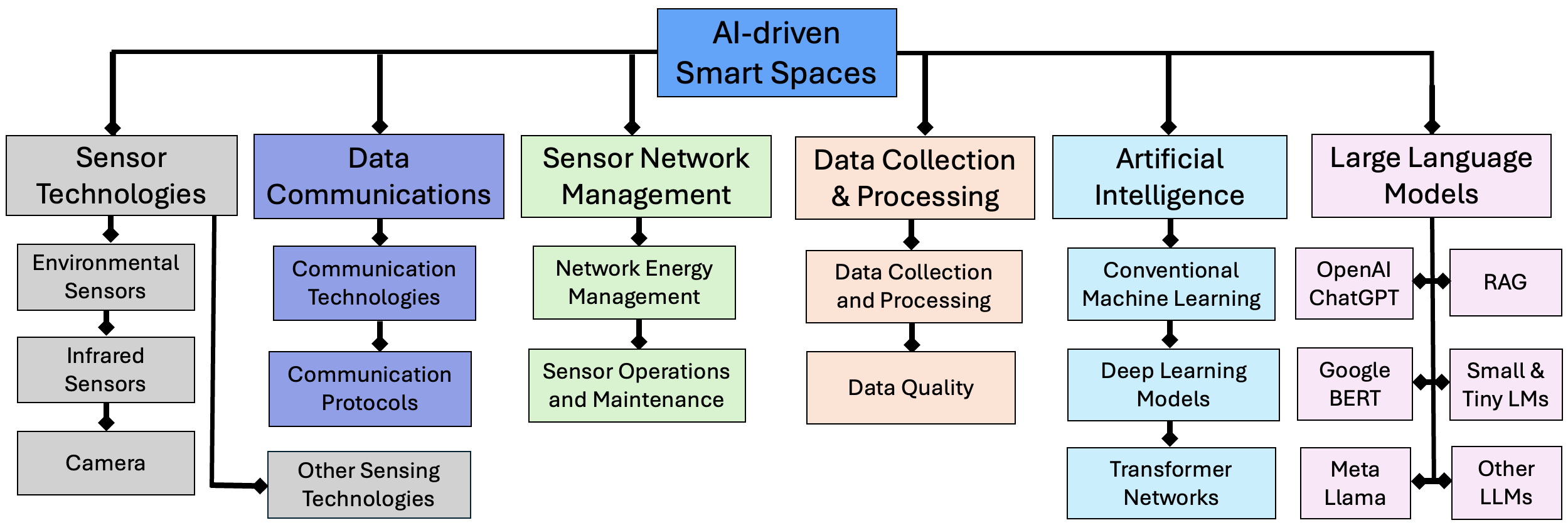}
  \caption{The scope and structure of the survey. }
  \label{fig:Scope}
\end{figure*}

The creation of AI-driven smart spaces requires the integration of sensing technologies and the establishment of communication networks within these spaces. Consequently, we structured this survey to align with the common requirements of IoT architecture. The core components, including the technologies and methodologies that underpin AI-driven smart spaces, are outlined in Figure~\ref{fig:Scope}. Each section presents these components, addressing the relevant technologies and methodologies in detail. To show the significance, we conclude each section with a dedicated subsection summarizing the associated challenges and limitations of the existing approaches.

In Section~\ref{sec:SensorDevices}, we discuss sensor technologies, which form the foundational components of smart spaces. Section~\ref{sec:Data Communications} explores the diverse range of communication technologies and protocols necessary for sensor data communication. Section~\ref{sec:Management} focuses on sensor network management, including energy optimization and operational reliability to ensure an effective and sustainable sensor network. 
In Section~\ref{sec:Data_Collect_Process}, we address the processes of collecting and processing data generated by sensors deployed in smart spaces. To emphasize the role of AI in enhancing AI-driven smart spaces, we first present the capabilities of conventional ML and DL models in enabling various applications. Following this, we introduce transformer networks and highlight LLMs as emerging learning techniques with transformative potential for developing AI-driven smart spaces. 
Given the importance of AI as a central focus of this article, we address its applications in two dedicated sections to maintain clarity and conciseness. In Section~\ref{sec:AI}, we provide a survey of smart space applications powered by ML, DL, and transformer networks. Then, Section~\ref{sec:LLMs} introduces the LLM families and models, emphasizing their effectiveness in advancing AI-based applications within smart spaces. 
Finally, Section~\ref{sec:Conclusion} concludes the paper with key remarks and future research directions, outlining opportunities for further advancements in AI-driven smart spaces.

\section{Sensor Technologies}
%=====================
\label{sec:SensorDevices}
Sensors are essential components in the development of smart spaces. They can collect various types of data from environments, providing insights into the state of the surroundings. Examples of such data include light, noise, temperature, humidity, pressure, motion, images, and other environmental factors~\cite{sivanathan2018classifying}. 
Generally, sensors are equipped with wired or wireless communication protocols such as Bluetooth, Wi-Fi, or LoRaWAN, enabling them to collect and transmit data to servers for storage and processing. In this section, we present sensor technologies that are suitable for creating smart spaces.

\subsection{Environmental sensors}
Environmental sensors are essential technologies used indoors to monitor and understand various conditions and aspects of the environment. Below are examples of sensors commonly utilized in indoor settings.

\textbf{Air quality sensors:} These sensors are vital technologies that monitor and provide essential data on various air pollutants. Examples of these pollutants include Carbon Monoxide (CO), Carbon Dioxide (CO$_2$), Nitrogen Dioxide (NO$_2$), Ozone (O$_3$), Formaldehyde (HCHO), Volatile Organic Compounds (VOC), and particulate matter like PM$_{2.5}$~\cite{baqer2023development, bousiotis2023monitoring, lopes2020indoor, motlagh2019indoor}. In smart spaces, these sensors enable real-time air quality monitoring, ensuring a clean, healthy, and comfortable indoor atmosphere. They can also trigger early warnings when air quality deviates from safe levels. The data gathered by these sensors can be used to intelligently control ventilation systems, adjusting airflow rates when pollution levels exceed safety thresholds~\cite{kong2022hvac}.

\textbf{Temperature and humidity sensors:} Temperature sensors play a crucial role in maintaining optimal thermal conditions. In indoor environments, they are used to assess thermal comfort and provide insights into personalized comfort experiences~\cite{morresi2024measuring}, \cite{aryal2020thermal}, \cite{li2020heat}, \cite{feng2023alert}. By integrating temperature sensors with physiological data, occupant-responsive models~\cite{fan2024data} can be developed to capture personalized thermal preferences and predict individual comfort levels more effectively. These temperature measurements enable HVAC systems to precisely adjust heating and cooling, ensuring occupant comfort while enhancing thermal satisfaction, e.g., through the integration of temperature and humidity sensing with digital twins to offer personalized thermal comforts~\cite{ma2024co}. Furthermore, automating HVAC systems based on real-time data from temperature sensors reduces excessive energy consumption, optimizing energy use within the space. Alongside temperature sensing, humidity sensors are also important sensing technology that enables maintaining proper indoor conditions by regulating relative humidity. These sensors help prevent mold growth, material damage, and discomfort caused by excess moisture or dryness, thereby creating a healthier and more comfortable indoor environment~\cite{bousiotis2023monitoring}. Consequently, the use of temperature and humidity sensors is essential in smart spaces to enhance personalized occupant experiences, improve well-being, and improve energy efficiency.

\textbf{Light and noise sensors:} The use of ambient light sensors (also known as photometric sensors) and noise sensors is crucial in indoor environments, as they help monitor and regulate light and noise levels, ensuring both visual and auditory comfort~\cite{hallett2018inexpensive, picaut2020low}. Indoor illumination is closely tied to the activities of the occupants~\cite{kumar2018sensing}. By sensing indoor light conditions, lighting systems can intelligently adjust and dim light levels to meet the visual comfort needs of individuals, optimizing energy use while maintaining comfort~\cite{dong2019review, motlagh2018iot}. 
Similarly, noise sensors are vital for detecting sound levels, which can be used to monitor occupancy or identify specific events occurring within indoor spaces~\cite{dissanayake2018detecting}. These technologies contribute to creating responsive and adaptive environments that enhance the overall well-being of occupants.

\subsection{Infrared sensors} 
Infrared sensors are among the most commonly used sensors in buildings, employed to detect the presence of people and monitor space occupancy, identify occupant activities and space utilization, and support energy management systems~\cite{alsafery2023sensing}. 

\textbf{Passive infrared (PIR) sensor:} This sensing technology, commonly known as a PIR (Passive Infrared) sensor, is widely used indoors to detect motion by passively sensing infrared (heat) radiation emitted by objects within its field of view~\cite{motlagh2019indoor}. PIR sensors operate in three modes: idle mode, where no motion is detected (i.e., no infrared radiation) and the sensor remains inactive; detection mode, which occurs when motion is detected; and triggering mode, when the sensor activates an event, such as triggering an alarm system to detect unauthorized entry. In smart environments, PIR sensors provide valuable data that can be used to enhance various comfort, health, and safety applications~\cite{wu2018low}. For instance, they enable motion-activated lighting, where sensors automatically turn lights on or off, and automated appliance control, where sensors manage electrical systems such as fans, smart curtains, heaters, or HVAC systems. By adjusting or turning off electrical systems when no motion is detected, these PIR-enabled applications contribute to significant energy savings~\cite{perra2021monitoring}.

\textbf{Ultralow-resolution infrared (IR) sensors:} This sensor technology is a type of thermal sensor that captures infrared (IR) radiation at a very low resolution. These IR sensors detect radiation emitted by objects and generate information using a limited number of pixels, restricting their ability to provide high accuracy or detailed outputs, such as precise temperature mapping~\cite{fasolino2024object}. 
However, these limitations make IR sensors cost-effective, energy-efficient, and privacy-preserving, making them well-suited for use in various smart environment applications. These applications typically involve basic infrared detection tasks, such as motion and presence detection, occupancy monitoring, proximity sensing, and simple gesture recognition~\cite{xie2023efficient, raghavachari2015comparative, xie2022energy}.

\subsection{Cameras} 
Cameras are among commonly used technologies in smart spaces. They offer a rich source of information from both users and systems, enabling the development of various applications such as security, access control, and activity recognition.

\textbf{Surveillance cameras:} These technologies also known as video surveillance systems, have become ubiquitous and widely accessible. Depending on the specific application and the desired image quality, commercially available cameras vary in frame rates, resolutions, and fields of view~\cite{tsakanikas2018video}. In smart environments, these cameras can be integrated with other sensor technologies and actuators, facilitating advanced applications such as facial recognition, gesture detection, augmented reality, and occupancy monitoring~\cite{elharrouss2021review}. However, despite their capabilities, surveillance cameras pose significant challenges related to privacy~\cite{adams2015future}, as well as data storage and processing~\cite{shao2017smart}. Therefore, careful consideration is required when designing and deploying these systems in smart environments to mitigate potential risks.

\textbf{Thermal cameras}: These technologies utilize a passive sensing approach, capturing infrared radiation (heat) emitted by objects~\cite{gade2014thermal}. While thermal cameras typically have lower resolution compared to surveillance cameras, they are well-suited for various privacy-preserving applications, such as people detection, counting, and tracking~\cite{altay2022use}. In smart environments, thermal cameras can be employed to monitor energy consumption of systems such as TVs, computers, lighting, and HVAC systems, helping optimize their operation. Additionally, these technologies can be used indoors to monitor occupancy in a privacy-conscious manner~\cite{naser2022multiple}.

\subsection{Other sensing technologies}
A wide variety of sensor technologies are available today, each tailored to measure specific variables and serve numerous applications~\cite{javaid2021sensors}. In this subsection, we present a few common sensor technologies that can be used to create smart spaces.

\textbf{Smart meters:} These devices are designed to measure, record, and monitor the consumption of resources such as electricity, gas, and water. In buildings, smart meters can be used to detect leaks or abnormal resource usage~\cite{hsia2020smart, cetin2017smart}. Therefore, in smart environments, they can play an important role in collecting consumption data, enabling a better understanding of space utilization. This data can be used to identify the energy consumed by specific systems (e.g., lighting, TVs, HVAC), and to develop models and forecasting methods for scheduling and optimizing resource utilization~\cite{nabavi2021deep}.

\textbf{Geophone sensors:} These highly sensitive devices can detect ground vibrations with high precision, making them widely used in seismic monitoring. Their accuracy and reliability also make them valuable for applications such as monitoring structural vibrations and movements in buildings~\cite{tang2017internet}. In smart environments, geophone sensors can be utilized for several purposes: (i) detecting occupancy to improve energy management, security, and space optimization, (ii) tracking occupant positions, providing insights into movement patterns, (iii) identifying footsteps and estimating direction, and enabling personalized tracking and behavior analysis, and (iv) analyzing space utilization to enhance interior design by adjusting layouts or seating arrangements.  

\textbf{Wireless sensing:} This sensing technique leverages the channel state information  
of wireless signals from communication technologies such as IEEE 802.11 (Wi-Fi) or cellular systems. By analyzing fluctuations in the radio waves transmitted between devices, this technique enables detecting the activities and events. Wireless sensing enables estimation of persons' presence, activity, location, motion, and their gestures~\cite{liu2019wireless, savazzi2016device, lin2020locater}. In smart environments, common devices such as wireless access points, TVs, mobile phones, laptops, and other smart devices—equipped with technologies like Wi-Fi, Bluetooth, or cellular networks—can facilitate the implementation of wireless sensing. Smart environments can harness wireless sensing for a wide range of applications, including security, health and wellness monitoring, emergency response, and intelligent interactions~\cite{wang2018device, khalili2020wi}.

\textbf{Ultrasonic sensing:} This technology employs inaudible sound waves, typically around 20 kHz, to detect and interpret interactions within an environment. By emitting acoustic signals through built-in speakers and capturing the reflected waves via microphones, this technique leverages the Doppler effect to analyze changes in frequency caused by moving objects. For instance, the research  in~\cite{lian2023echosensor} presents EchoSensor, a sensing system that utilizes ultrasonic sensors for intrusion detection in smart homes. EchoSensor system employs existing audio devices to transmit and capture ultrasonic signals, processing the reflections to identify patterns of human gait cycles. In addition, this research addresses that ML methodologies and signal processing allow to differentiate between authorized individuals and potential intruders, pinpointing the potentials of ultrasonic sensing for enhancing smart home security.

\textbf{Actuators:} These technologies are designed to convert energy—whether electrical, hydraulic, or pneumatic—into mechanical motion. Actuators function as devices or processes that translate control signals into physical movement. They are employed in a wide range of applications, from industrial automation to consumer electronics, enabling the operation of various systems~\cite{arshi2023advancements}. In smart environments, actuators play a crucial role in optimizing energy efficiency. For instance, they control the operation of HVAC and lighting systems, regulating airflow or adjusting lighting levels based on environmental conditions. They can also be integrated with smart curtains to modulate light exposure, contributing to energy conservation and enhanced user comfort.

\textbf{Smartphones and wearables:} 
Equipped with advanced sensor technologies, smartphones and wearables generate valuable data that can seamlessly integrates into smart spaces. Smartphones, for instance, provide GPS, accelerometer, and Bluetooth data, enabling precise tracking, indoor positioning, and an understanding of the services users require within the space~\cite{torres2021data}. 
An example of their application is proximity-based contact tracing, where smartphones use Bluetooth and wireless systems to detect interactions between users. Utilizing these built-in sensors, smartphones can be used for public health measures such as disease tracking in indoor environments~\cite{ranjha2022aect}. 
Wearables, with their physiological and biomedical sensors, allow continuous monitoring of users' health, safety, and activities, such as tracking health metrics within smart spaces~\cite{dian2020wearables}. Furthermore, these devices can support privacy-aware solutions, ensuring that smart spaces respect user confidentiality while leveraging their capabilities~\cite{seng2023machine}.

\subsection{Challenges and limitations} 
Commercially available sensor technologies are affordable and provide a wide range of environmental data from the locations where they are deployed. Their low cost and availability make it feasible to deploy them in large numbers, enabling the creation of smart environments. However, despite these advantages, they come with the following limitations:

\textbf{Sensing capabilities:} Low-cost sensors typically have a limited number of sensor units. Some sensors are designed to measure a specific variable, such as temperature, while others can measure multiple variables on a single sensor board, such as temperature, relative humidity, and CO$_2$. In reality, there are many variables that can be measured to fully understand the state of an indoor environment, but not all sensors are equipped to capture all of these variables. For example, some sensors can measure temperature, relative humidity, and CO$_2$, while others measure variables such as temperature, noise, light, and PM$_{2.5}$. To overcome this limitation, virtual sensors powered by AI and machine learning can be developed. These virtual sensors estimate unmeasured variables by using available data as a proxy~\cite{zaidan2023virtual}. For instance, PM$_{2.5}$ and temperature measurements can be used to estimate CO$_2$ and black carbon concentrations, providing a more comprehensive understanding of indoor environmental conditions~\cite{liu2024estimating}.

\textbf{Communication capabilities:} 
Low-cost sensors, when not using cables for communication, typically employ at least one type of wireless communication technology. Each technology has its own set of characteristics, such as connectivity robustness, bandwidth, coverage, and energy consumption. For instance, while one low-cost sensor might offer Bluetooth connectivity, another might provide both Bluetooth and Wi-Fi options. Bluetooth is known to have robustness connectivity concerns~\cite{bronzi2016bluetooth}, while Wi-Fi might be more demanding in terms of energy consumption~\cite{putra2017comparison}, which is a crucial consideration for battery-powered sensors. Therefore, when planning the connectivity and communication strategy for sensors, it is necessary to consider the availability and features of each communication technology. 

\textbf{Sensing accuracy:} Low-cost sensors often experience reduced sensing accuracy over time. Although these sensors are initially calibrated in a controlled laboratory environment, their measurement precision tends to degrade, leading to measurement drifts. However, in the smart environments, AI and machine learning-based calibration techniques can be employed to re-calibrate these sensors more frequently~\cite{zaidan2022intelligent}, helping to ensure accurate sensing over time.

\textbf{Energy consumption:} Most low-cost sensors are battery-powered and have limited energy resources. To create smart environments and improve the management of sensors deployed in these settings, it is essential to plan an effective data transmission strategy. While some sensors generate small amounts of data that require less power to transmit, the frequency of data transmission can be adjusted based on the importance of the measured variables. This helps to extend the battery life. For sensor devices that capture images and require more bandwidth and transmission power, mechanisms can be implemented to selectively transmit a limited number of images, ensuring the extension of battery life~\cite{newell2019review}.

\textbf{Security and privacy:} In smart environments, the use of cameras and other sensor devices that can potentially reveal individuals' identities is often known as privacy-intrusive, raising significant security and privacy concerns. To address these issues and protect individuals' privacy, a range of privacy-preserving techniques can be employed, including encryption, anonymization, access control, and differential privacy~\cite{zhang2017security}.

\section{Data Communications}
%=====================
\label{sec:Data Communications}
In this section, we present the most common wireless communication technologies and protocols that can be used to establish smart spaces.

\subsection{Communication technologies}
The diverse range of technologies used to establish smart spaces—such as sensors, actuators, smart phones, smart wearables, and smart TVs—each have unique communication requirements that vary based on factors like deployment location, lifespan constraints, and data traffic patterns. Therefore, selecting the right communication technology for a smart space is a critical design decision, as it can greatly influence the performance, power consumption, and overall functionality of the connected devices, ultimately affecting the efficiency of the smart environment. This section provides an overview of the wireless  technologies suitable for smart spaces.

\textbf{Bluetooth Low Energy (BLE)} is a short-range wireless communication technology specifically designed for Internet of Things (IoT) applications, serving as an extension of classic Bluetooth. BLE is cost-effective to implement and supports a typical range of up to 100 meters. Operating in the 2.4 GHz ISM band, BLE can achieve data rates of up to 2 Mbps, especially with BLE v5.0, which offers high-throughput communication. Optimized for low power consumption, BLE is ideal for battery-operated devices~\cite{Bluetooth}. It is a highly versatile and widely adopted technology, commonly used in IoT-enabled devices such as consumer electronics. BLE provides reliable solutions for indoor localization, proximity detection, and energy-efficient operation~\cite{shan2022indoor}, making it a an efficient technology to be used for creation of smart environments.

\textbf{Wi-Fi}, also known as WLAN, encompasses a variety of standards tailored to support both short-range and long-range scenarios, each designed for specific use cases. Among these standards, 802.11ac and 802.11ax (Wi-Fi 6 or HaLoW) are particularly suitable for smart indoor environments. The 802.11ac standard provides a theoretical throughput of up to 6.7 Gbps, making it ideal for high-bandwidth applications. It typically offers a range of up to 45 meters indoors and 90 meters outdoors. In contrast, the 802.11ax standard presents an indoor range of 70 meters and an outdoor range of 250 meters, along with a theoretical throughput of up to 10 Gbps. This enhanced capacity makes it an excellent solution for large-scale sensor deployments and big data applications~\cite{wifialliance}. Due to its support for high-data-rate IoT applications, Wi-Fi requires significant power, especially in scenarios that demand continuous, plugged-in operation. Despite being a dominant technology, in smart environments, Wi-Fi particularly can be utilized applications demanding significant power and high data rates, such as video steaming scenarios~\cite{zhou2020optimization}. 

\textbf{Long Range Wide Area Network (LoRaWAN)}, commonly referred to as LoRa, is a low-cost, low-power technology designed to support a maximum data rate of 50 kbps. This technology targets a battery life of up to 10 years, making it particularly well-suited for resource-constrained IoT devices. LoRaWAN operates on a star-of-stars topology and does not facilitate device-to-device communication. The technology employs an adaptive data rate technique, optimizing transmission parameters to enhance both energy efficiency and network longevity~\cite{loraalliance}. With minimal energy consumption, LoRaWAN can maintain reliable connectivity over long distances while providing high accuracy. This capability enables effective remote monitoring and control of IoT devices. These features, combined with its low-power requirements, make LoRaWAN exceptionally suitable for smart space applications~\cite{ahsan2021smart}.

\textbf{Low Power Wireless Personal Area Networks (6LoWPAN)} is a low-power, short-range communication technology specifically designed for transmitting IPv6 packets. It offers a maximum data transfer rate of 250 kbps and supports a range of up to 200 meters. 6LoWPAN facilitates the deployment of mesh network topologies and implements the IEEE 802.15.4 standard for device connectivity~\cite{culler20096lowpan}. In smart environments, the features of 6LoWPAN make it an ideal technology for interconnecting various smart devices, including sensors, actuators, and home appliances~\cite{sanila2020performance}.

\textbf{ZigBee} is a low-power, cost-effective, and low-data-rate wireless technology based on the IEEE 802.15.4 standard, designed for short-range communication. Operating in a mesh network configuration, ZigBee supports up to 65,536 devices and offers a maximum data transfer rate of 250 kbps~\cite{zohourian2023iot}. This technology allows devices in close proximity to communicate directly without the need for routers or access points. With a battery life of up to 20 years, ZigBee devices typically have a communication range of 10 to 100 meters, depending on environmental factors and device configurations. Due to its energy efficiency and flexible network scalability, ZigBee is widely used in smart environments, particularly in home automation systems, such as smart sensors, lighting, door locks, and home controllers~\cite{akestoridis2020zigator}.

\textbf{Z-Wave} is a wireless technology designed for low-power, low-data-rate applications, with a range of up to 100 meters in open air and up to 20 meters indoors. It supports multiple data rates, including 100 kbps for maximum throughput, and 9.6 kbps or 40 kbps for more energy-efficient communication. Z-Wave allows bidirectional communication, facilitating a robust mesh network topology. In theory, a Z-Wave network can accommodate up to 232 devices, though for optimal performance and reliability, a practical limit of around 50 connections is recommended~\cite{z-wavealliance}. Due to its reliability and ease of integration, Z-Wave is widely adopted in smart home automation and security systems~\cite{vattheuer2023z}.

\textbf{Thread} is a versatile wireless technology built on the IEEE 802.15.4 standard, enabling IPv6-based wireless mesh networking. It is optimized for low-power operation, making it ideal for battery-powered devices. Thread supports a range of up to 30 meters in open air and up to 15 meters indoors~\cite{ThreadGroup}. Initially, this technology is intended for smart home environments, enabling various applications such as appliances, access control, climate control, smart thermostats, lighting, as well as safety and security systems, including cameras and alarms~\cite{lan2018latency}.

\textbf{Radio Frequency Identification (RFID)} is a communication technology used for identifying and tracking objects. The ISO/IEC 18000 series establishes standards for RFID, including information management protocols. For example, ISO/IEC 18000-6B allows simultaneous reading of up to 10 tags with a large data storage capacity, while ISO/IEC 18000-6C enables the reading of hundreds of tags at once, although with lower individual data storage per tag. RFID is a cost-effective solution, offering a typical range from a few centimeters to several meters. This makes it ideal for efficient tracking systems, especially in indoor positioning and localization scenarios. RFID-based indoor localization is particularly useful in smart environments for monitoring daily activities and tracking objects or visitors~\cite{lubna2022radio}.

\textbf{Near Field Communication (NFC)} is a subset of RFID technology, primarily used for close-range applications. It enables communication within a range of up to 2 centimeters and supports data transfer rates from 46 kbps to 1.7 Mbps~\cite{NFC}. NFC can enhance indoor navigation systems by allowing users to update their location by simply tapping NFC tags placed throughout an environment, effectively addressing the limitations of GPS in indoor settings~\cite{ozdenizci2015nfc}. The short operational range of NFC, combined with its ease of use and security features, makes it an ideal choice for secure identification systems. For instance, it can be used for access control in smart environments, enabling users to authenticate themselves when entering restricted areas.

\textbf{Extended Coverage – GSM – Internet of Things (EC-GSM-IoT)} is a cellular wireless technology developed specifically to support IoT applications that require low data rates and extended coverage. By utilizing existing 2G infrastructure, EC-GSM-IoT enables efficient and reliable IoT connectivity over a broader GSM coverage area, making it ideal for smart environments where numerous devices require secure, energy-efficient connections~\cite{liberg2017cellular}. This low-power technology is designed to achieve up to 10 years of battery life using just a 5 Wh battery. While EC-GSM-IoT does not support voice communication, it offers flexible data rates ranging from 350 bps to 70 kbps, adjusting based on coverage and transmission requirements. These characteristics make it an excellent choice for automating and controlling key home systems, such as HVAC, lighting, and surveillance cameras~\cite{obeidat2021remotely}.

\textbf{Long Term Evolution for Machines (LTE-M)} is an open-standard cellular IoT technology designed to support a wide range of low to mid-range IoT applications. LTE-M offers several advantages, including low device costs, extended coverage, and long battery life, while maintaining the capacity to support a high density of devices per cell. With data rates reaching up to 384 kbps, LTE-M is well-suited for applications requiring higher data throughput and low latency~\cite{liberg2017cellular}. A key benefit of LTE-M is its support for mobility, voice, and SMS, which makes it highly adaptable across various smart home applications. In smart environments, LTE-M can power numerous systems such as home security, closed-circuit television (CCTV), and fire alarm networks, where reliable connectivity and real-time responsiveness are essential~\cite{joe2017band}.

\textbf{Narrowband Internet of Things (NB-IoT)} is a low-power cellular IoT standard designed to provide a 20 dB gain over GPRS, making it particularly effective for indoor applications. With a target battery life of up to 10 years and a data rate of up to 100 kbps, NB-IoT can support up to 50,000 IoT devices per cell~\cite{liberg2017cellular}. These features make it an ideal choice for smart environments. In such settings, NB-IoT can be used for various applications, including smart lighting systems, indoor environment monitoring, home automation and control, as well as enabling efficient, accurate, and low-latency indoor positioning~\cite{xue2019indoor}.

\subsection{Communication protocols}
The communication protocols are a set of rules that govern the transmission and exchange of data across sensor networks. These protocols typically follow either the publish-subscribe model, the request-response model, or capable to implement both of the models. 

\textbf{Publish–Subscribe and Request–Response Methods:} 
In the publish–subscribe model, communication between data producers (publishers) and consumers (subscribers) is decoupled. Publishers send data to a central broker, which then distributes it to subscribers based on predefined topics or criteria. This model is particularly effective for systems requiring many-to-many communication, as it reduces the overhead of direct communication between devices. In contrast, the request–response model involves direct communication between devices. A client sends a request to a server, and the server responds with the required information. This model is well-suited for scenarios that demand immediate, one-to-one interactions~\cite{dizdarevic2019survey}. 
Both models offer distinct advantages. In the publish–subscribe model, publishers and subscribers do not need to be aware of each other's presence. The complexity of connections is reduced by eliminating point-to-point communication, allowing a single subscriber to receive data from multiple publishers, and a single publisher to broadcast data to multiple subscribers. Additionally, publishers and subscribers do not need to be active simultaneously, as the broker can queue and store messages, forwarding them when clients become active. On the other hand, the request–response model has the advantage in providing reliable, real-time interactions, particularly when the server can manage high traffic volumes and meet client demands effectively~\cite{sachs2010benchmarking}. 
When designing smart environments, the choice between the communication models depends on the infrastructure, type of sensors, scale of sensor deployment, and the computational capabilities of the system.

\textbf{Message Queuing Telemetry Transport (MQTT)} is a lightweight messaging protocol supported by the Advancement of Structured Information Standards (OASIS), a nonprofit standardization consortium. It operates on a publish-subscribe model and, due to its minimal header design, is highly efficient for devices with limited resources, such as low-power sensors. This makes it an ideal choice for sensor networks with low bandwidth and high latency. MQTT relies on the TCP transport protocol, ensuring reliable communication. Since MQTT is designed to be lightweight, data is exchanged in plain text by default, therefore to secure data transmission, it uses TLS/SSL encryption~\cite{banks2014mqtt}. In MQTT, two key entities are involved in communication: clients and brokers. Clients can either publish messages or subscribe to receive messages. The broker, acting as the central component, receives messages from publishing clients, filters them by topic, and forwards them to the appropriate subscribers. The broker can also store messages for new subscribers or discard them if no subscribers exist for a given topic~\cite{yudidharma2023systematic}.

\textbf{Constrained Application Protocol (CoAP)} is a lightweight messaging protocol supported by the Internet Engineering Task Force (IETF), an open standards organization~\cite{shelby2014constrained}. CoAP follows a request-response model and runs over the UDP transport protocol, offering minimal overhead, making it ideal for communication in constrained environments. The protocol architecture consists of two layers: a messaging layer and a request-response layer. The messaging layer manages message transmission and ensures reliability, while the request-response layer handles the actual exchange of data~\cite{bansal2023enhancing}. When a CoAP client sends a confirmable (CON) request to a CoAP server, the server generates an acknowledgment (ACK) in response. The response data is embedded within the ACK message, a process known as a piggybacked response. To secure communication, CoAP uses Datagram Transport Layer Security (DTLS), running over UDP. Additionally, CoAP includes a feature that allows clients to continuously receive updates on a requested resource from the server, enhancing the traditional request-response model. The IETF further extends CoAP's capabilities to support a publish-subscribe model, enabling even more flexible communication approach~\cite{koster2019publish}.

\textbf{Advanced Message Queuing Protocol (AMQP)} is an open-standard communication protocol supported by the OASIS standard. AMQP is designed to ensure interoperability across a broad spectrum of devices, enabling efficient data exchange between platforms developed in different programming languages, particularly suited for heterogeneous systems. The protocol operates using both the publish–subscribe and request–response models and runs on top of the TCP transport protocol. AMQP incorporates a store-and-forward mechanism to enhance message reliability. AMQP also implements TLS/SSL protocols to ensuring the confidentiality and integrity of data, meeting the security requirements~\cite{standard2012oasis}. Moreover, while less commonly used compared to MQTT and CoAP, AMQP offers robust messaging capabilities at the price of higher resource demand~\cite{yudidharma2023systematic}. AMQP is typically employed for communication between software applications, servers, and hosts in smart home environments~\cite{adiono2017intelligent}.

\textbf{Data Distribution Service (DDS)} is a protocol based on the decentralized publish-subscribe model that is standardized by the Object Management Group (OMG)~\cite{OMG2015}. It operates without the need for a broker component, enabling asynchronous data exchange between publishers and subscribers via a data bus. By eliminating the need for message brokers, DDS enables devices to join or leave the network at any time. 
DDS is particularly well-suited for real-time systems, utilizing both UDP and TCP transport protocols to provide extensive QoS capabilities. For security, DDS employs TLS, DTLS, and DDS Security (DDS-Sec) protocols, ensuring robust data protection. To address the high overhead of TLS and DTLS protocols in constrained environments, the OMG DDS security specification introduces a comprehensive security model, with a service plugin interface (SPI) architecture, allowing security implementations in IoT systems~\cite{saleh4872730publish}.

\textbf{HTTP} is a widely used application-layer protocol governed by IETF standards. Over the years, different versions of HTTP have been developed, each designed to enhance the protocol's performance. HTTP follows a request–response model and relies on TCP as its transport protocol, ensuring reliable data transmission. It commonly uses JSON for data exchange, and for security purposes, it employs TLS encryption, which is often referred to as HTTPS~\cite{fielding1999hypertext}. 
Despite its widespread use, HTTP faces limitations in IoT environments due to its complexity, large header sizes, and high power consumption.
Additionally, HTTP is often associated with REST, a set of guidelines that prescribes using specific HTTP methods (e.g., GET, POST, PUT, DELETE). RESTful HTTP (or REST over HTTP) supports CRUD (Create, Retrieve, Update, Delete) operations, enabling the development of web applications and simplifying state management for IoT nodes~\cite{severance2015roy}. 
In environments where communication and power efficiency are less critical, such as edge and cloud computing systems, RESTful HTTP remains a viable option~\cite{dizdarevic2019survey}.

\textbf{Extensible Messaging and Presence Protocol (XMPP)} is a communication protocoldefined by the IETF which is designed for near-real-time data exchange between network entities. It supports various communication models, including request-response, publish-subscribe, end-to-end, and multicast interactions. XMPP is built on Extensible Markup Language (XML) and operates within a distributed client-server architecture. Data is exchanged asynchronously using XML fragments, enabling functionalities like messaging, presence updates, and request-response interactions~\cite{saint2011extensible}. For security, XMPP employs Transport Layer Security (TLS) for authentication and uses the Simple Authentication and Security Layer (SASL) for encryption. Its scalable design allows for custom protocol extensions, making it adaptable to a wide range of applications, including instant messaging, IoT deployments, and multi-user conferencing.

\textbf{QUIC} is a recently developed protocol which is standardized by the IETF, designed to improve upon traditional TCP-based protocols including HTTP/1.1 and HTTP/2, particularly the latency, connection establishment, and transport layer security. Unlike TCP, QUIC operates over UDP and follows a request-response architecture similar to HTTP~\cite{QUICprotocol}. 
QUIC enhances web performance by integrating key features such as multiplexing, encryption, congestion control, and connection migration directly into the transport layer. These optimizations make it especially effective in mobile and unreliable network environments. Additionally, QUIC improves energy efficiency by enabling faster data transmission, reducing the time devices spend actively communicating, which in turn lowers power consumption. Its advanced features make QUIC an ideal choice for latency-sensitive applications, real-time services, and IoT deployments, where reliable and efficient data transmission is essential~\cite{iyengar2021quic}.

\subsection{Challenges and limitations}
The communication technologies and protocols that enable data transmission in smart environments come with their own set of challenges and limitations, some of which we will briefly address here.

\textbf{Communication Technologies:} Most short-range wireless communication technologies are designed for low-power, low data-rate, and constrained IoT devices, which introduces significant challenges when creating smart environments. Below, we highlight a few examples. BLE faces difficulties in maintaining stable connections in dynamic channel environments, leading to performance degradation. When a connection is lost, nodes must restart the recovery process, which involves exchanging multiple control packets and results in increased power consumption~\cite{lee2017cable}. Zigbee operates in the same frequency bands as WLAN, which increases the likelihood of interference in wireless environments. This interference leads to packet transmission delays and reduced reliability, necessitating adaptive cooperation mechanisms between Zigbee and WLAN to mitigate these issues~\cite{hong2015performance}. 

Wi-Fi, despite its widespread availability and robust connectivity, is power-hungry in IoT environments. Designed primarily to optimize bandwidth, range, and throughput, Wi-Fi is not well-suited for power-constrained applications, particularly those relying on battery-powered devices~\cite{mesquita2018assessing}. LoRa (Long Range), although specifically designed for low-power, low-data-rate communication, operates in unlicensed frequency bands. This can lead to an increase in packet collisions due to shared spectrum use~\cite{temim2020enhanced}. Additionally, as the number of connected devices in a LoRa network grows, contention and interference increase, ultimately degrading network performance~\cite{premsankar2020optimal}. 

Short-range communication technologies also face security threats and vulnerabilities~\cite{nkuba2023zmad}. For instance, Z-Wave has been shown to be susceptible to cyberattacks. By exploiting crafted data, attackers can potentially disable key features of Z-Wave-enabled IoT devices. One significant vulnerability arises during wireless firmware updates, where attackers can remotely take control of Z-Wave device operations~\cite{kim2020s}. Other example technologies that are often vulnerable to security threats include RFID and NFC. RFID tags can carry sensitive personal information, making them a target for exploitation by adversaries~\cite{ahmed2021cloud}. Similarly, while NFC operates within a few centimeters and requires physical device contact—such as tapping devices together to exchange personal information—this approach, though secure in proximity, can be less convenient compared to alternatives like app-based controls.

\textbf{Communication Protocols:} Although highly effective in enabling data exchange within IoT networks, various communication protocols present distinct challenges and limitations. For instance, HTTP faces difficulties in IoT environments due to its complexity, the large header size of TCP, and its high power consumption\cite{severance2015roy}. XMPP, on the other hand, relies on XML, leading to larger message sizes, which are inefficient for bandwidth-constrained networks. Additionally, XMPP's dependence on a persistent TCP connection, combined with its lack of efficient binary encoding, makes it unsuitable for lossy, low-power wireless networks commonly found in IoT environments. In terms of security, while XMPP supports basic SASL and TLS protocols, it lacks advanced native features like end-to-end encryption~\cite{celesti2017enabling}.

Other protocols, such as MQTT and CoAP, also have security vulnerabilities. Although MQTT is an efficient protocol for IoT, it is susceptible to various cyberattacks because it was originally designed for trusted IoT networks~\cite{kotak2019comparative}. Basic security measures such as username/password authentication and SSL/TLS encryption are often inadequate, necessitating additional protective measures~\cite{toe2023lightweight}. Similarly, CoAP, which operates over UDP, lacks a handshake mechanism, making it vulnerable to IP spoofing attacks. This vulnerability can escalate into more severe threats, such as Distributed Denial-of-Service (DDoS) and amplification attacks, within IoT environments~\cite{ray2021daiss}.

Additionally, the emerging QUIC transport layer protocol comes with its own limitations. One of its key features, 0-RTT (zero-round-trip time), reduces latency but introduces the risk of replay attacks. Since QUIC operates over UDP, it is also susceptible to amplification attacks, where attackers exploit the absence of initial handshake verification to overwhelm servers. Furthermore, adoption challenges arise due to compatibility issues with legacy network infrastructure, as some firewalls and middleboxes may not be optimized for handling UDP traffic efficiently. While QUIC encrypts most packet contents for security, certain metadata remains unencrypted to facilitate routing, raising concerns about potential privacy risks through traffic analysis~\cite{joarder2024exploring}.

\section{Sensor Network Management}
%=====================
\label{sec:Management}

Creating smart environments requires the effective management of IoT networks to efficiently control their operations. This involves meeting several key requirements, where we present in this section.

\subsection{Sensor network energy management}
Efficient energy management is crucial for the design and deployment of sensors in smart spaces. Since many IoT devices are battery-powered and often installed in hard-to-reach locations, such as ceilings, it is vital to implement strategies that minimize energy consumption while maintaining long-term operational efficiency~\cite{zanaj2021energy}. Sensors equipped with high-power wireless communication modules are particularly significant energy consumers, which highlights the need for optimization techniques to extend their operational lifespan. For battery-powered sensors that require regular recharging or replacement, frequent maintenance is often impractical and costly~\cite{jayakumar2014powering}. Moreover, scaling the number of sensors increases the complexity of the network, demanding more efficient coordination and dynamic energy optimization techniques. For instance, sensor signals are prone to interference, as many devices operate on the same frequency bands. As the network scales, additional energy may be required for signal re-transmission due to lost or corrupted data. To ensure stable and optimal energy consumption and performance as the network expands, current state-of-the-art research introduces scalable procedures to address these challenges~\cite{mansour2023internet}.

\textbf{Energy Harvesting:} Sensor devices that frequently transmit small amounts of data can utilize energy harvesting techniques—by converting ambient energy sources such as solar, vibration, or thermal energy into electrical power—to reduce dependence on batteries or grid power. For instance, research has proposed battery-free, light-based sensor systems powered by lighting, enabling energy harvesting even in indoor environments~\cite{landivar2024batteryless}. Similarly, other studies have introduced dual-use energy harvesting systems that leverage both light and radio frequency technologies to power data transmission from environmental sensors~\cite{orfanidis2019towards, katz2024towards}. Furthermore, another research  employed reinforcement learning to optimize operation of energy-harvesting sensor devices, and dynamically configure the sensors operation based on environmental conditions. Employing solar panels and super-capacitors for energy storage, proposed system in the research adapts to varying light levels, adjusting its duty cycle to conserve energy during low-light periods and increase data collection when the energy is abundant~\cite{fraternali2020aces}. 
Indeed, implementing energy harvesting methodologies in smart spaces is therefore a sustainable solution, significantly reducing energy requirements and operational costs within the sensor network. By minimizing the need for battery replacements or recharging, energy harvesting also enhances the long-term efficiency and scalability of IoT deployments, contributing to more resilient and cost-effective smart environments.

\textbf{Sensor Energy Management Techniques:} Several research studies have proposed various energy management strategies to optimize sensor power usage. One commonly employed technique is \textit{duty cycling}, which conserves energy by periodically switching sensors on and off during periods of inactivity. Many sensor systems utilize heavy duty cycling, with over 90\% of their operational time spent in low-power sleep modes, interrupted by brief bursts of activity~\cite{jayakumar2014powering}. The ML can further enhance duty cycling, improving energy efficiency by up to 30\% without compromising sensing accuracy~\cite{belapurkar2018building}.
Another approach is \textit{power gating}, which selectively powers down individual components or modules of the sensors when they are not in use. This method is particularly effective for sensors with multiple functional modules, as these often do not need to operate simultaneously, leading to significant energy savings~\cite{jayakumar2014powering}. 
\textit{Sleep scheduling} is another energy-saving technique that puts sensors into low-power modes during inactivity and activates them only when necessary. This is especially useful in periodic data collection scenarios where continuous sensor operation is not required~\cite{callebaut2021art}. \textit{Dynamic Voltage Scaling (DVS)} is a technique that dynamically adjusts the voltage supplied to sensors based on real-time demands, significantly extending sensor node lifespan. DVS also adapts to changes in workload and network conditions, ensuring sensors only consume the required amount of power. A task-driven feedback DVS algorithm can further optimize energy usage by dynamically scaling voltage while using feedback loops to correct errors~\cite{tuming2010dynamic}. 
In addition to these methods, a variety of other strategies can extend sensor battery life. These include selecting low-energy components, optimizing hardware design, and employing techniques such as \enquote{race to sleep} and \enquote{think before you talk}~\cite{callebaut2021art}.

\textbf{Energy-Aware Communication Protocols:} In sensor networks, energy-efficient communication protocols are crucial for minimizing power consumption while ensuring reliable data transmission. The protocols Low-Power Wireless Personal Area Networks (LoWPANs) and Low-Power Short-Area Networks (LPSANs)—such as RFID, NFC, Zigbee, Bluetooth, Z-Wave, and 6LoWPAN—are specifically designed for short-range communication with minimal energy consumption. These energy-aware protocols achieve efficiency by optimizing transmission frequency, allowing sensors to preserve data integrity while minimizing power usage~\cite{zanaj2021energy}. Additionally, Low-Power Wide-Area Networks (LPWANs) are designed for long-range communication with broad coverage, maintaining low power consumption. Examples of LPWAN technologies include LoRa, Sigfox, NB-IoT, LTE-M, and EC-GSM-IoT, which are ideal for applications that require extensive network reach with minimal energy use~\cite{boulogeorgos2016low}. In large-scale sensor deployments, LPWANs enable extended sensing and monitoring over vast areas while ensuring energy efficiency.

\textbf{Other Approaches:} Several additional techniques in the literature focus on optimizing energy consumption in IoT sensor networks. Among the most effective are \textit{ML-based energy management} approaches, which enhance sensor energy efficiency by dynamically optimizing sampling rates. These techniques can predict the energy consumption of various sensors in smart environments and adjust their sampling rates in real time, based on factors like occupancy patterns. By leveraging historical data and ML algorithms, these systems enable more intelligent decision-making. For example, in a smart space, during periods of low occupancy, ML models can instruct sensors to reduce their sampling rates, conserving energy without compromising sensing accuracy~\cite{bereketeab2024energy}. Another approach is \textit{sensor fusion}, which integrates data from multiple sensors to improve accuracy while reducing the energy consumption of individual devices. By combining information from various types of sensors—such as temperature, humidity, and motion—sensor fusion allows for more efficient data collection and processing, ultimately conserving power of sensor devices~\cite{tsanousa2023fusion}.

\subsection{Sensors Operations and Maintenance}
Effective sensor operations and management are crucial for the seamless functioning of smart spaces. To ensure accurate and reliable data collection, the following key aspects must be addressed:

\textbf{Maintenance and Real-Time Monitoring:} In smart spaces, regular maintenance—such as cleaning or replacing sensors—is crucial to ensure sensor longevity and prevent performance degradation. Real-time monitoring and condition-based maintenance of sensors and networking infrastructure enable the identification of inefficiencies and the adjustment of sensor operations only when necessary, ensuring optimal performance while minimizing the costs of physical inspections~\cite{martins2023online}. 
Real-time monitoring in smart spaces often involves occupancy detection sensors, such as Passive Infrared (PIR) motion sensors, which help differentiate between high- and low-occupancy zones~\cite{laidi2018udeploy}. The combined use of PIR motion detectors and CO$_2$ sensors can further estimate occupancy levels and analyze patterns within a space, providing valuable insights for improved environmental control~\cite{motlagh2021monitoring}. Additionally, digital twins play a key role in sensor deployment and management by creating virtual replicas of physical spaces. These digital models allow for the simulation of real-world conditions, offering insights through descriptive, diagnostic, predictive, and prescriptive analytics, enhancing decision-making and sensor maintenance~\cite{motlagh2023digital}.

\textbf{Sensor Fault Detection:} Sensors are susceptible to issues such as drift, bias, or complete failure, and early detection of these faults is critical for maintaining reliable sensor performance~\cite{zaidan2022dense}. Research has proposed various distributed fault detection methods to identify both permanent and intermittent sensor faults. For example, establishing trust relationships between sensors in smart spaces can effectively detect sensor faults while using minimal computational resources~\cite{guclu2016distributed}. To ensure the overall reliability of a sensor network, it is essential to regularly assess the performance of sensors to verify that they produce accurate measurements. This can be achieved through frequent testing, statistical comparisons of measured variables, and correlation analyses between sensor data and that of an accurate portable sensor within the smart space. Sensors that exhibit drift and produce anomalous data patterns are referred to as \enquote{anomaly sensors}. These anomalies can be identified through outlier analysis and by calibrating the sensor's measurements against accurate reference sensor. Sensor failure—where a sensor stops transmitting data—is a common issue, often caused by faults in the power unit, sensing components, or communication modules. Continuous real-time monitoring and regular inspections enables preventing sensor failures~\cite{duche2013sensor}.

\subsection{Challenges and Limitations}

\textbf{Sensor Calibration:} Low-cost sensors, which form the backbone of smart space sensing infrastructures, often experience degradation in sensing quality over time, even under normal operating conditions~\cite{lagerspetz2019megasense}. As a result, regular calibration is crucial to maintain sensor accuracy and ensure the reliability of collected data. Calibration involves establishing a precise correlation between the sensor's raw output and the actual measured value~\cite{aula2022evaluation}. In practice, this process is typically conducted either in a laboratory setting or by comparing the measurements of sensors to highly reliable and precise reference sensors~\cite{motlagh2020toward}. To ensure accurate and consistent sensor measurements, calibration is necessary before sensors are deployed in real-world environments~\cite{motlagh2020low}. In smart spaces, where sensors are deployed at large scales, regular calibration is required. However, removing and re-installing individual sensors for laboratory calibration is impractical. Therefore, in-situ calibration—where sensors are calibrated in their deployed locations—becomes essential~\cite{delaine2019situ}. To achieve this, there is a need to develop opportunistic calibration mechanisms, such as using a recently calibrated sensor to calibrate others within the smart space.

\textbf{Data Security:} Sensor networks are highly vulnerable to various cyber threats, including eavesdropping, data tampering, and denial of service (DoS) attacks. Ensuring robust security involves implementing encryption, authentication, and secure communication protocols to protect the integrity of both data and the network. Due to their limited computational capabilities, sensors are particularly susceptible to attacks such as node replication and eavesdropping. While symmetric-key cryptography can be employed to enhance security, it introduces additional challenges due to resource constraints~\cite{burhanuddin2018review}. In smart spaces, sensors often collect crucial and sensitive information about individuals, making cyber attacks a significant threat to the collected data. Without adequate security mechanisms, intruders can easily compromise sensors and networking infrastructure, such as Wi-Fi routers, to access this data~\cite{ray2020iot}. Therefore, implementing an effective security strategy to counter cyber attacks and safeguard data collected by sensors is a critical challenge~\cite{fernandes2016security}.

\section{Data Collection, Processing, and Quality}
%=====================
\label{sec:Data_Collect_Process}

\subsection{Data Collection and Processing}
The creation of smart environments relies on deploying heterogeneous sensor technologies, generating, data with differing volumes, velocities, varieties, veracity, value, and vulnerabilities~\cite{wongthongtham2017big}. 
For instance, an environmental sensor designed to measure temperature and air quality information may transmit data every few seconds. In contrast, cameras—whether infrared or surveillance—may operate continuously, streaming high volumes of image data at a rapid pace~\cite{wongthongtham2017big}. As a result, deploying diverse sensor technologies in smart spaces leads to the generation of complex datasets with varying time-stamped data.
In addition, the sensor heterogeneity and varying standards often result in data being produced in different formats. For example, environmental sensors might output semi-structured data such as \textit{.json} or \textit{.xml}, while cameras and microphones generate unstructured data like video and audio streams, respectively~\cite{azad2020role}. These variations increase data complexity, necessitating flexible data collection and handling strategies tailored to smart environments. A solution however would be using specialized time-series databases designed for these types of sensor deployments.  

In smart environments, the collected sensor data can be analyzed in real-time to provide services to occupants or used for advanced analytics to enhance the environment's functionality. The continuous data streams from multiple sensors generate large volumes of data, often referred to as \enquote{big data} which demand significant computational resources—especially when applying AI and machine learning (AI/ML) models to deliver AI-based services. To offer AI/ML-based services, several data processing functions must be executed. These include data preprocessing, which involves noise removal, data conversion, normalization, and labeling; feature extraction; classification; and the execution of specific ML models~\cite{mahdavinejad2018machine}. Each of these functions requires fast and accurate data processing to ensure the delivery of real-time services to users.

Edge computing and cloud computing both offer effective solutions to meet the data storage and processing demands of smart environments~\cite{hayyolalam2021edge}. The choice between these computing approaches depends on the specific design of the smart space, requiring careful consideration of the respective advantages and disadvantages of each. Naturally, cloud computing offers substantial computational power but raises concerns regarding latency, privacy, and bandwidth. In contrast, edge computing places the computing facilities closer to where sensor devices are located, which helps improve privacy and bandwidth concerns and reduce latency, enabling real-time services~\cite{pan2017future}.

In smart environments, edge computing can serve as the main computing facility, connecting sensors, actuators to collect and process data, enhancing the performance and efficiency of these environments, by delivering timely and accurate data analytics and decision-making at the network's edge. 
Edge computing can also leverage machine learning and deep learning techniques, facilitate communication and coordination among devices, including user-connected devices such as smartphones, and adapt to dynamic and complex settings~\cite{bhatia_overview_2022}. For instance, it can enable the implementation of models that learn from user behaviors and the deployment of LLM-based models to provide customized and personalized services for the users of the environment.

The cloud computing on the other hand can function as a processing platform either by directly receiving data from sensors or by serving as a combined processing facility alongside edge computing, following the edge-cloud computing continuum architecture~\cite{kokkonen2022autonomy}. While cloud computing may introduce some latency and privacy concerns, it provides substantial storage capacity and facilitates the execution of AI/ML models that require large datasets, thereby enhancing the quality and functionality of services within smart environments~\cite{cao2021survey}.

\subsection{Data Quality}\label{sec:Data Quality}
The quality of sensor data is fundamental to the accurate interpretation of environmental events in smart environments. Data quality, both in terms of its qualitative and quantitative aspects, is assessed based on key criteria such as accuracy, completeness, validity, consistency, uniqueness, and timeliness. In smart environments, the integrity of sensor data is crucial, as it directly influences decision-making processes. However, several factors, such as sensor degradation, measurement drift, network failures, and battery depletion (in battery-powered devices), can lead to deteriorating data quality, resulting in incomplete, inaccurate, inconsistent, noisy, outdated, or redundant data streams~\cite{mukhopadhyay2021artificial}. 

Sensor technologies that are suitable to be deployed at indoor environments, typically designed to be low-cost, often have limitations in terms of sensing accuracy. Fortunately, recent advancements in artificial intelligence (AI) and machine learning (ML) have enabled the development of effective calibration models that significantly enhance the accuracy of these sensors~\cite{cheng2019ict}. For example, state-of-the-art ML-based methods, including regression models, ensemble techniques, and neural networks, have been applied to improve the measurement accuracy of key environmental variables such as particulate matter (PM2.5) and carbon dioxide (CO2) levels—both critical for indoor air quality monitoring~\cite{aula2022evaluation, motlagh2019indoor, zaidan2020intelligent}. Furthermore, specialized ML models, such as Long Short-Term Memory (LSTM) networks, have been employed to process sequential and temporal data (e.g., air quality measurements)~\cite{zaidan2020intelligent}. Convolutional Neural Networks (CNNs) are commonly used for image and video analysis, while Recurrent Neural Networks (RNNs) and Autoencoders (AEs) are leveraged for tasks such as data compression and reconstruction. Variational Autoencoders (VAEs) offer additional benefits for data modeling, and Generative Adversarial Networks (GANs) assist with data augmentation and synthesis. In addition, Deep Reinforcement Learning (DRL) is utilized for control and decision-making within the environment~\cite{lakshmanna2022review}.

In addition to sensor calibration, the quality of sensor data can be affected by errors, outliers, anomalies, and missing data points. AI and ML-based techniques provide robust solutions for addressing these data quality challenges. For instance, data cleaning algorithms can detect and correct anomalies by removing or repairing invalid, missing, duplicate, or inconsistent data points~\cite{mukhopadhyay2021artificial}. Moreover, cross-labeling and alignment methods can enable the identification and correction of inconsistencies across multi-modal datasets ensuring reliability in the labeled data, resulting in improved data quality~\cite{zhang2023multisense}.

Data fusion techniques can also enhance the comprehensiveness and accuracy of environmental data by combining inputs from multiple sensors, thus enriching datasets that might otherwise be incomplete, noisy, or uncertain. Additionally, the application of data provenance—techniques that track and record the origin, history, and ownership of data—offers a way to further improve data quality. By attaching metadata that describes the source, context, reliability, and trustworthiness of the data, provenance systems provide additional layers of validation.

In nutshell, maintaining and improving sensor data quality in smart environments is essential for reliable decision-making. This can be achieved by implementing AI/ML-based data processing pipelines that enhance data quality at both the edge and cloud, where sensor data is collected and processed. By integrating advanced ML techniques, these pipelines can improve the accuracy and reliability of sensor data, which in turn optimizes the overall performance of smart environments.

\subsection{Challenges}

\textbf{Processing costs:} The use of AI/ML techniques enable the learning of complex, nonlinear patterns from high-dimensional and unstructured data, allowing for effective sensor calibration, anomaly detection, and handling of missing data. However, the adoption of AI/ML in smart environments presents the following challenges. The first challenge relates to the data availability. The lack of large, labeled datasets specific to smart environments hinders the training of accurate models. Generally, data in IoT systems is often sparse, unlabeled, or incomplete, making it difficult to apply supervised learning methods without significant pre-processing efforts. The second challenge relates to the computation and memory costs. AI/ML models, especially deep learning techniques, require substantial computational resources to execute and large memory to store data. The third challenge refers to the model interpretability. Many AI/ML models operate as "black boxes" providing predictions without offering insights into their decision-making process within the smart environments. Enhancing the transparency of these models remains an ongoing challenge.

\textbf{Privacy and security:} 
In smart environments, IoT systems continuously collect and process large volumes of data, which may contain sensitive information about individuals and their surroundings. This raises significant privacy and security concerns, particularly with respect to safeguarding personal data and complying with regulatory frameworks such as the General Data Protection Regulation (GDPR) and the California Consumer Privacy Act (CCPA)~\cite{carlson2020general}. Followings include some of the key privacy and security challenges in sensor data processing. 

The data generated by IoT sensors can potentially reveal individuals' identities and behaviors, making it critical to implement privacy-preserving approaches. For example, sensors deployed in smart environment inadvertently capture personal activities, leading to a increased risk of privacy violations. 
In addition, ensuring compliance with stringent data protection regulations is a challenge in smart environments. These regulations impose strict guidelines on how personal data should be collected, stored, and shared, necessitating advanced privacy-preserving techniques. 

To meet the security requirements, the state-of-the-art encryption techniques, such as the Advanced Encryption Standard (AES) and Transport Layer Security (TLS), are commonly used to secure data during transmission. Additionally, homomorphic encryption schemes, such as the Brakerski-Gentry-Vaikuntanathan (BGV) cryptosystem~\cite{alwarafy2020survey}, enable secure data aggregation without exposing the underlying information. However, while these methods enhance security, they also introduce computational overhead, making it challenging to implement them for the smart environments.

Moreover, robust data governance frameworks are essential to prevent unauthorized access and mitigate cyber threats. Inadequate access controls and improper data management strategies can result in data breaches, demanding the development of effective privacy-preserving algorithms.

\section{Artificial Intelligence}
%=====================
\label{sec:AI}

\begin{figure}[ht]
  \centering
\includegraphics[width=0.75\linewidth]{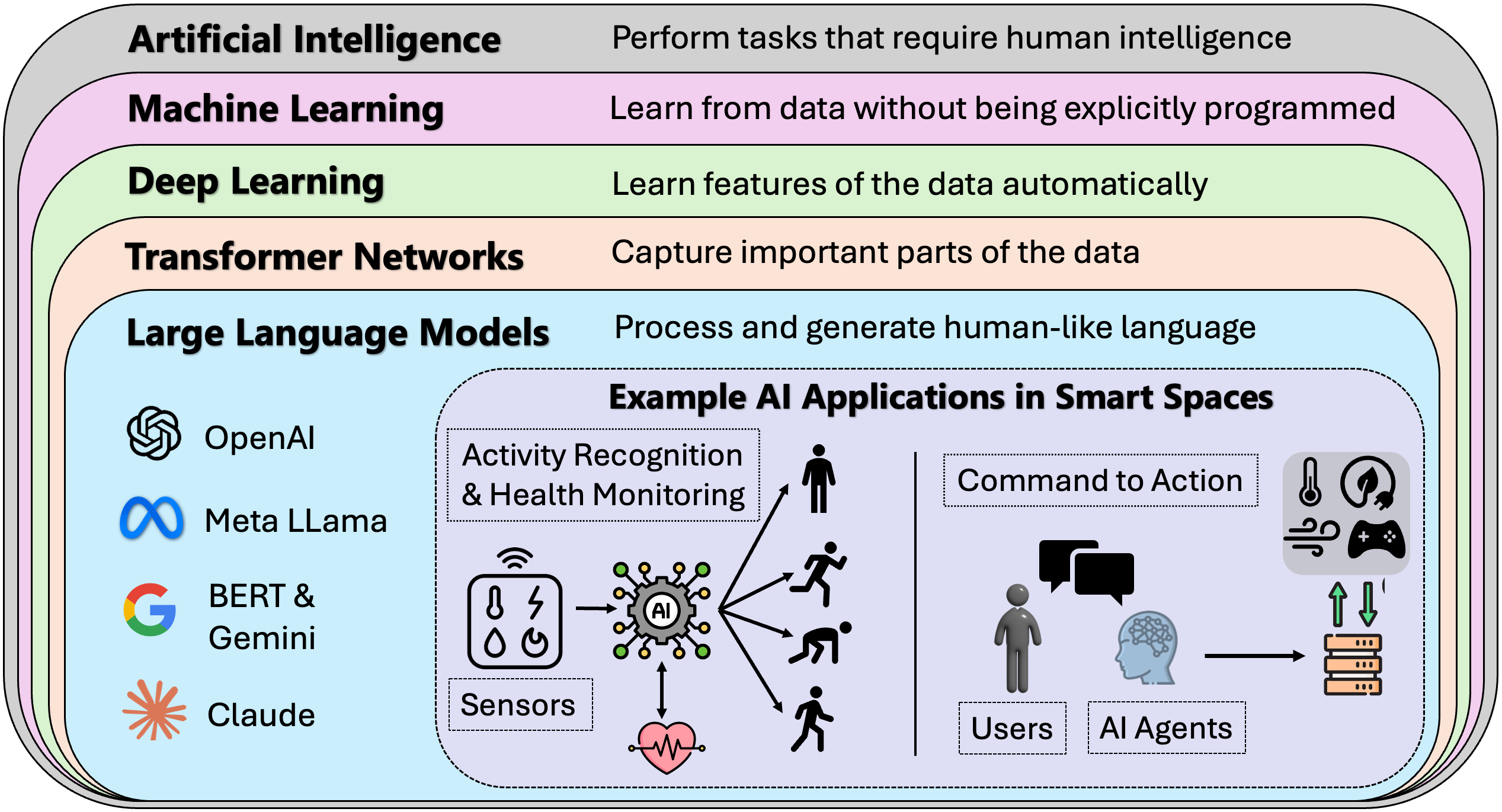}
  \caption{Advancing AI techniques from conventional ML methods to LLMs.}
  \label{fig:AI}
\end{figure}

The artificial intelligence (AI) and machine learning (ML) are terms often used interchangeably. However, while AI broadly refers to the ability of computers to perform tasks in real-world contexts, ML specifically involves the development of algorithms that enable systems to analyze data, recognize patterns, and make informed decisions. Creation of smart spaces relies heavily on the ML tools to provide AI-driven functionalities. This section explores the AI and ML techniques, needed to enable the capabilities of smart spaces. Figure~\ref{fig:AI} illustrates the progression of AL methodologies, tracing their development from conventional ML techniques to advanced transformer networks and large language models (LLMs). We use this framework to develop this section.

\subsection{Conventional machine learning models}
Conventional machine learning (ML) models use relatively simple structures of interconnected neurons to analyze and interpret data. These models are trained on historical datasets to develop predictive capabilities, enabling them to generalize and make accurate inferences on new data. Common examples of these models include perceptrons, multilayer perceptrons (MLPs), feed-forward artificial neural networks, random forests, logistic regression, naive Bayes, support vector machines (SVM), k-nearest neighbors (KNN), and extreme gradient boosting (XGBoost)~\cite{kelleher2020fundamentals}. A major strength of conventional ML models lies in their versatility for tasks such as classification, detection, regression, clustering, and pattern recognition, making them effective methods for human activity recognition in smart spaces.

For instance, the multilayer perceptron (MLP) model is a common neural network architecture, composed of an input layer, one or more hidden layers, and an output layer. Each neuron connects to every neuron in the subsequent layer, enabling the MLP to learn complex patterns through non-linear activation functions. By iteratively updating weights via backpropagation, the network can approximate any continuous function, achieving high classification accuracy. One study demonstrated that an MLP model with 256 neurons achieved 98\% activity classification accuracy when trained on wearable device data from sensors on the ankle and wrist~\cite{majidzadeh2021human}. 
NLPs, when combined with k-pattern algorithms, can further enhance the accuracy of activity classification and prediction~\cite{bourobou2015user}, enabling user-centric solutions in smart spaces.

Another effective model is the random forest, an ensemble learning algorithm used for both classification and regression. A random forest comprises multiple decision trees, each trained on a randomly selected subset of the data, and aggregates their outputs to produce a robust prediction. Random forest models have shown high accuracy in applications such as fall detection and activity recognition using human skeleton features~\cite{ramirez2021fall}. Random forest shows an accuracy of 91\% for people counting and 98.13\% for occupancy detection~\cite{violatto2020anomaly}. 
  
Additionally, other conventional ML models such as XGBoost, and SVMs have proven to be powerful for anomaly detection~\cite{alnaqbi2023anomaly}, which is essential for improving life quality through continuous monitoring in smart spaces. By identifying irregularities in daily activities, these models can detect early signs of health issues, particularly for the elderly and individuals with disabilities. 
 
Moreover, conventional ML models show efficiency in significantly reducing unnecessary electricity consumption within smart spaces. For example, ANNs with backpropagation and variable learning rate (VLR) can automate energy control by toggling electrical components on and off based on occupancy or use patterns~\cite{akhinov2021development}.

\subsection{Deep learning models}
Deep learning (DL) architectures enhance traditional machine learning (ML) approaches by incorporating complex models designed to handle large datasets. DL architectures consist of multiple hidden layers, allowing for automated feature extraction from raw data and enabling the processing of diverse unstructured data types, including images, text, speech, and video. Such capabilities make DL models highly effective for tasks such as image recognition, activity recognition, natural language processing, and autonomous systems, making them powerful tools for real-time data analysis in smart spaces. 
DL architectures are generally classified into supervised learning models, such as Convolutional Neural Networks (CNNs) and Recurrent Neural Networks (RNNs), and unsupervised learning models, such as Autoencoders (AEs).  
Based on these foundational architectures, numerous DL models have been developed, some of which we address in this subsection. 

CNN as a fundamental deep learning architecture, consists of three primary components including the convolutional layer, the pooling layer, and the fully connected layer. Convolutional layers extract features from input data, identifying essential patterns such as edges, textures, and shapes. Pooling layers downsample the feature maps, effectively reducing spatial dimensions and computational complexity. Finally, fully connected layers integrate the features learned across the network to enable predictions or classifications. 
In smart spaces, CNNs can facilitate enable privacy-preserving, real-time applications such as occupancy monitoring, activity recognition, and people counting~\cite{xie2023efficient}. For instance, YOLO (You Only Look Once) is a widely used CNN architecture designed for real-time object detection by combining object localization and classification tasks within a single convolutional network. The lightweight design of YOLO makes it an ideal choice for deployment in smart spaces~\cite{zhou2021deep}.

RNNs are another important DL architectures, specifically designed for processing sequential data. RNNs incorporate self-looping connections that enable them to retain a hidden state, capturing information from previous inputs. This unique feature makes RNNs well-suited for tasks where context and order are crucial, such as time series analysis, natural language processing, activity recognition, and speech recognition. In smart spaces, RNNs can offer a range of applications; for example, they can improve indoor localization by generating movement path data based on Wi-Fi fingerprints~\cite{shin2021movement}, and enable human activity recognition~\cite{park2018deep}.

Long Short-Term Memory (LSTM) networks are a specialized variant of RNNs designed to capture long-range dependencies in sequential data. The LSTM architecture enables these models to retain context over extended sequences, making them ideal for analyzing time-sensitive sensor data. By effectively capturing temporal dependencies, LSTM models can provide more accurate predictions. In smart spaces, LSTMs are particularly useful for applications such as human activity recognition~\cite{singh2017human} and energy consumption forecasting, supporting optimized energy management strategies~\cite{syed2021household}.

The Gated Recurrent Unit (GRU) network is an optimized variant of the RNN architecture, offering similar capabilities to LSTM networks for processing sequential data. With a simplified design and fewer parameters, GRUs improve computational efficiency while effectively capturing temporal dependencies. This makes them well-suited for tasks  such as time series analysis, natural language processing, and speech recognition, where modeling long-term dependencies is essential. As a result, GRUs are ideal for developing applications such as sensor-based activity recognition~\cite{lu2022multichannel}, air quality prediction~\cite{sarkar2022air}, and electricity consumption forecasting~\cite{jrhilifa2024forecasting}.

Autoencoders (AEs) are DL models that are developed for unsupervised learning, capable of capturing essential features of data while filtering out irrelevant information. These features makes AEs particularly effective for functions such as anomaly detection. Hence, in smart spaces, AEs can be employed to identify anomalies in time-series sensor data. For example, AEs can detect unusual behaviors such as irregular power consumption, control system failures, sensor malfunctions~\cite{dissem2024neural}, and anomalies at indoor air quality~\cite{wei2023lstm}.

\subsection{Transformer networks}
Transformer networks, a deep learning architecture innovation proposed by Google researchers, are distinguished by their attention mechanism, enabling a model to dynamically weigh the significance of elements within a sequence to capture complex contextual relationships~\cite{vaswani2017attention}. Unlike earlier models such as RNNs and LSTMs, which process data sequentially, transformers operate in parallel, enhancing computational efficiency and scalability. This parallelism allows transformer networks to outperform prior deep learning models in both accuracy and speed, making them particularly effective for tasks involving time-series data, natural language processing, and other applications requiring the modeling of complex and context-dependent relationships in data. Additionally, transformers demonstrate strong performance even in supervised tasks with limited labeled data, a valuable feature for various smart space applications~\cite{chen2022transformer}.

Compared to traditional deep learning models like convolutional neural networks (CNNs), LSTMs, and support vector machines (SVMs), transformer networks exhibit superior performance in user behavior and activity recognition in indoor environments~\cite{huang2023human, chen2022transformer}. This advantage arises from the self-attention mechanism of transformers, which enables the modeling of spatial-temporal dependencies, enhancing classification and activity detection accuracy~\cite{xiao2022two, mao2024spatio}. Furthermore, transformers excel in forecasting, allowing for predictive applications such as estimating users' next activities and their durations, a critical capability for enabling context-aware services in smart spaces that offer personalized and anticipatory interactions~\cite{roy2021action, kim2021tap}. Beyond activity detection, transformers facilitate indoor localization and mobility pattern recognition, enabling user classification, location prediction, and schedule estimation~\cite{zhang2023hierarchical}. Such insights can foster healthier environments, particularly benefiting elderly individuals, patients, children, and people with special needs by monitoring mobility and behavior in smart spaces~\cite{bader2023using, jovan2023multimodal}. 

Due to their long-range sequence modeling capabilities, transformers are increasingly employed in indoor activity detection using WiFi signal strength and coverage, for example, in modeling indoor human localization and mobility through WiFi connectivity patterns~\cite{chen2024aiot, trivedi2021wifimod}. Transformers utilize self-attention to process WiFi channel state information (CSI), achieving high indoor localization accuracy~\cite{zhang2022tips}. This method also offers a privacy-preserving alternative to camera-based systems by reducing reliance on visual data for pose estimation~\cite{zhou2023metafi++}. Notably, transformers facilitate privacy-preserving techniques that mitigate privacy concerns—a critical requirement for smart spaces~\cite{lee2023transformer, sheng2024cdfi}. 

Other studies leverage wearable-based sensing, incorporating sensors like inertial measurement units (IMUs)~\cite{xiao2022two}, or utilize smartphone-based approaches, applying Vision Transformers (ViT) for image processing in human localization tasks~\cite{gufran2023vital}. RFID-based methods, which measure received signal strength indicator (RSSI) values, are also used for activities like fall detection for elderly individuals indoors~\cite{khan2024tag}.

Transformer networks contribute to energy efficiency and operational optimization in smart spaces. For instance, transformers can optimize HVAC systems and forecast energy demands by using models such as the Temporal Fusion Transformer, which captures long-range dependencies in sequential energy data through attention mechanisms~\cite{zhou2024interpretable}. These models can deliver accurate energy load predictions based solely on smart meter readings~\cite{wang2022transformer}. Furthermore, multi-task learning frameworks based on transformers, utilizing environmental variables, support short-term multi-load energy forecasting~\cite{ji2023multi, cen2024multi}. Integrating transformer-based, context-aware systems with deep reinforcement learning (DRL) enables multi-zone HVAC control optimization, balancing energy consumption with thermal comfort, thereby enhancing quality of life in smart environments~\cite{deng2022toward}. 

Transformers also excel in transfer learning, enabling knowledge transfer across tasks to improve model performance in dynamic environments~\cite{akbari2022transformer}. In smart spaces, transformer networks can adapt to user preferences and behaviors, supporting user-centric interactions and assistance~\cite{tiwari2023inbuilt}. They even facilitate immersive metaverse interactions, where users can control and engage with their surroundings through avatars in virtual environments~\cite{zhou2023metafi++}.

To further enhance transformer performance, recent studies have explored integrating the self-attention mechanism into other neural network types, improving the model’s ability to focus on relevant features in the input data. Hybrid models that combine transformers with RNNs or CNNs utilize attention mechanisms to filter out irrelevant information, boosting prediction accuracy for human activity recognition in smart spaces~\cite{tang2022dual}. These hybrid approaches provide a flexible and powerful framework for solving complex problems, significantly enhancing smart space applications.

\subsection{Challenges and Limitations}

\textbf{Implementation Complexity:} In smart spaces, the effectiveness of DL and ML models relies heavily on data collected from various sensors, followed by a training phase to perform specific tasks. Conventional ML models typically require supervised learning, necessitating extensive labeled datasets and data preprocessing—such as outlier detection, data imputation, and feature selection—to ensure data integrity and model accuracy. Due to noise and inconsistent data formats, preprocessing may also involve filtering and scaling techniques~\cite{bzai2022machine}. Selecting and configuring models for smart spaces is challenging, as DL/ML models like CNNs and RNNs need tuning to accommodate memory and processing limitations of the computing devices. In addition, these models must be optimized for low latency and energy efficiency, often requiring specialized edge computing configurations and hardware expertise~\cite{abdel2021deep}. To address these constraints, strategies such as model compression and efficient memory management are essential, as limited resources can impact the performance and scalability of conventional ML models, complicating deployment in resource-constrained computing environments~\cite{dalal2020analysing}. Architectures such as on-device computing, edge server-based solutions, and hybrid systems combining edge and cloud resources need careful selection based on specific use cases and device limitations. Achieving optimal performance on resource-limited hardware therefore demands substantial expertise in both ML and embedded systems~\cite{merenda2020edge}. As a result, deploying and managing DL/ML models in resource-constrained environments is challenging and often requires specialized expertise due to the complexity of model training and optimization.

\textbf{Data Privacy and Security:} Deploying AI in smart spaces raises significant privacy challenges, as these models often process sensitive user data, including identity, health information, personal habits, and preferences. Since smart spaces inherently involve local data collection, on-device processing and localized data handling are essential for mitigating privacy risks. Existing studies proposes methods such as Federated Learning (FL)—which enables model training across decentralized devices, such as IoT nodes, without sharing raw data—to address the privacy challenges of IoT deployments. By training AI/ML models directly on local devices without transmitting raw data, FL helps preserve user privacy. However, FL remains vulnerable to \enquote{poisoning} attacks, in which training data or model parameters are deliberately manipulated to mislead the model~\cite{motlagh2024population}. Despite this vulnerability, FL is particularly valuable in smart indoor spaces where sensitive location data is involved, as it reduces data transmission risks while maintaining model performance~\cite{wu2022prediction, patel2024trustable}. On the security side, many IoT devices within smart environments are resource-constrained, which limits the use of complex security protocols without impacting performance. This limitation highlights the need for lightweight, efficient security solutions that balance strong protection with operational efficiency~\cite{badar2023secure}. In summary, preserving data privacy and developing robust security approaches are central challenges for smart spaces, where safeguarding users' information is essential.

\section{Large Language Models}
%=====================
\label{sec:LLMs}

Large language models (LLMs) -- that are deep learning models trained on big datasets -- are gaining the research momentum. With billions of parameters, LLMs can capture complex linguistic patterns and structures, and leverage the self-attention mechanisms and scalability of transformer architectures to process and understand languages across extensive sequences. This makes LLMs invaluable tools for various natural language processing (NLP) tasks such as text classification, sentiment analysis, and machine translation~\cite{raiaan2024review, an2024iot}.

Integrating LLMs within smart spaces enables a wide range of applications that enhance operational performance, comfort, and promote healthier living environments. For instance, in smart spaces, LLMs can promote sustainable behaviors and improve user interactions by dynamically adjusting environmental settings and minimizing energy waste through predictive analytics~\cite{giudici2023assessing}. LLMs can facilitate user-centric interactions, offering assistance with functions such as pose estimation and activity recognition~\cite{li2023chatiot}. Interestingly, in multilingual environments, LLMs support communication by providing user-friendly notifications in various languages~\cite{simunec2023smart}.
LLMs can also interact directly with the physical world through sensors deployed in the spaces, inferring user activities from sensor data to improve contextual awareness~\cite{xu2024penetrative, zhong2024casit}. Additionally, LLMs can be utilized to develop recommendation systems that provide personalized suggestions—optimizing for instance workspace setups, promoting energy-saving habits, and guiding users to suitable areas based on behavioral patterns—all of which contribute to sustainable practices~\cite{wu2024survey, lubos2024llm}.

\subsection{Retrieval-Augmented Generation (RAG)} 
Applying LLMs to real-world applications often requires integrating Retrieval-Augmented Generation (RAG) techniques, which equip generative AI models with information retrieval capabilities. RAG enables LLMs to access external data sources in real-time without additional training. This process involves two main stages including a \enquote{Retrieval Stage}, where relevant documents or snippets are fetched from an external corpus based on the input query, and a \enquote{Generation Stage}, where the LLM synthesizes information from the retrieved content to generate contextually accurate responses. This dual-stage process allows RAG-equipped LLMs to provide precise, real-time, and contextually enriched responses.

Presently, most studies in the literature apply RAG techniques to tasks like open-domain question answering and document retrieval. For example, the study in~\cite{ma2024fine} explores RAG-equipped models \enquote{RepLLaMA and RankLLaMA} for retrieval tasks using a multi-stage pipeline. Whereas, RepLLaMA identifies relevant documents from a large corpus, and RankLLaMA re-ranks these based on relevance scores to prioritize the most contextually appropriate documents. Another study ~\cite{an2024iot} enhances LLM reasoning in IoT applications by incorporating RAG to retrieve domain-specific IoT knowledge, thereby facilitating complex reasoning tasks.

In the literature, research on LLMs and RAG techniques applied to smart spaces is currently limited. In smart spaces, RAG-equipped LLMs can significantly improve tasks such as human activity recognition, anomaly detection, WiFi-based human sensing, and indoor localization by up to 65\%~\cite{an2024iot}. Therefore, using LLMs with RAG capabilities can enhance the functionality of smart spaces, leading to more interactive and user-centered experiences. 
In the followings, we address well-known LLM models that can be utilized to enhance the capabilities of smart spaces.

\subsection{OpenAI's Generative Pre-trained Transformer (GPT)}
The GPT is a state-of-the-art language model and a pioneering framework in generative artificial intelligence. Developed by OpenAI, GPT leverages advanced deep learning techniques. Since the release of its first version, GPT-1, in June 2018, the model has steadily evolved, with each new version demonstrating enhanced abilities in capturing language patterns and generating coherent, contextually relevant text. These capabilities provide opportunities to use GPTs to enhance the functionalities of smart spaces. 

The GPT-1 model was trained using unsupervised learning with a 12-layer transformer architecture, where each layer included self-attention and feed-forward networks. It employed 117 million parameters and was trained on a dataset of over 7,000 unpublished books~\cite{radford2018improving}.

The second evolution, GPT-2, was released with significantly increased capacity, consisting of 1.5 billion parameters and trained on a 40GB dataset of internet text. The GPT-2 architecture scaled up to 48 transformer layers, enabling it to handle tasks such as language translation, summarization, and question-answering without task-specific fine-tuning. In smart environments, GPT-2 can be applied to sensor event sequence prediction and real-time activity recognition of space users. This GPT-2-based approach predicts future sensor events in an autoregressive manner, outperforming traditional LSTM-based methods and providing anticipatory support in smart spaces~\cite{takeda2023sensor}. 

GPT-3 was later introduced with 175 billion parameters and trained on a dataset of approximately 570 GB of internet text. With 96 transformer layers, GPT-3 was designed to support few-shot and zero-shot learning capabilities. Building on this foundation, the GPT-3.5 version was subsequently released, offering enhanced performance through optimized training techniques, including Reinforcement Learning from Human Feedback (RLHF)~\cite{brown2020language}. 
Existing literature demonstrates how GPT-3 enhances smart spaces by enabling context-aware and adaptable responses, surpassing the limitations of traditional rule-based systems. For example, GPT-3 can translate open-ended commands, such as \enquote{get ready for a party}, into actionable device controls such as configuring lights and playing music. This is achieved through prompt engineering, where GPT-3 converts natural language commands into structured JSON outputs, which are then processed to control smart space devices~\cite{king2023get}. 
In addition, similar to earlier versions, GPT-3.5 can be leveraged for zero-shot activity identification in smart spaces by utilizing sensor-based activity monitoring. For instance, the study in~\cite{civitarese2024large} applies GPT-3.5 to interpret environmental sensor data, transforming raw sensor inputs into descriptive text that enables GPT-3.5 to classify activities based on its pre-trained knowledge. Another study~\cite{saleh2024follow} introduces \enquote{Follow-Me AI}, a GPT-3.5-powered system integrated with centralized AI agents that gathers both user preferences and environmental data to enhance user experiences in smart spaces. Follow-Me AI is designed to enable real-time adjustments to temperature, lighting, and occupancy, aligning with user preferences while also optimizing energy efficiency and maintaining data privacy.

Released in 2023, GPT-4 further enhanced GPT's reasoning and generalization abilities through training on a diverse dataset that included internet text, books, and specialized sources, allowing it to handle complex, domain-specific tasks more effectively~\cite{achiam2023gpt}. GPT-4 introduced multimodal input capabilities, processing both text and images. Its input is divided into smaller units, called tokens, which are analyzed by Transformer layers. With a context length of up to 32,000 tokens, GPT-4 can predict subsequent tokens based on learned patterns to generate human-like responses~\cite{devi2024quantitative}. 
Building on GPT-4, two new versions were introduced: GPT-4 Turbo, optimized for faster response times and lower computational demands, and GPT-4o, an omni model capable of processing and generating multiple media types, including text, audio, images, and video. GPT-4o is particularly notable for its rapid response to audio inputs, with an average latency comparable to human response time (around 320 ms). Key features of GPT-4o include web access, multimodal data processing, and enhanced code and mathematical capabilities, which together support improved reasoning abilities~\cite{hurst2024gpt}.

Integrating GPT-4 with smart spaces can offer advanced features and support context-aware, real-time decision-making in these environments, bridging the gap between digital reasoning and physical-world interactions. 
For example, the use of ChatGPT-4 have the potential to understand natural language commands and translate them into functional code, enabling home automation systems to activate code generated directly from user input~\cite{andrao2024sounds}. This capability democratizes smart environments by allowing users to control devices and systems through natural language, enhancing accessibility and user experience. 
The study in~\cite{xu2024penetrative} introduces \enquote{Penetrative AI}, which leverages GPT-4 to interpret smartphone sensor data and infer user activities, such as walking or remaining stationary, both indoors and outdoors. This system achieves over 90\% accuracy in distinguishing between indoor and outdoor movements. Another study~\cite{king2024sasha} presents a smart assistant named \enquote{Sasha}, which uses GPT-4 to respond to commands with creative, goal-oriented action plans. For example, when given the command \enquote{make it cozy}, Sasha interprets this abstract request into actionable JSON-based plans, adjusting smart devices such as lighting and temperature to create a comfortable environment. 
The research in~\cite{ahn2023alternative}, GPT-4 is applied to autonomously control HVAC systems in an office building. Integrated with a building simulation model, GPT-4 receives real-time indoor and outdoor CO$_2$ data and energy consumption metrics via a Python-based co-simulation interface. Based on this information, GPT-4 determines optimal control actions for variables for instance damper positions and chilled water temperature, aiming to minimize energy consumption while keeping indoor CO$_2$ levels below 1000 ppm. This example underscores the potential of large language models as effective decision-making agents, leveraging pre-trained general knowledge to perform complex, domain-specific tasks.

\subsection{Meta AI’s Large Language Model (LLaMA)} 
The LLaMA is a powerful family of autoregressive language models designed to provide efficient, high-quality language understanding for both general and specialized applications. Since the release of its first version in February 2023, Meta has developed several versions, including models with up to 65.2 billion parameters. LLaMA models are based on a transformer architecture and are pre-trained on a mixture of publicly available data sources, including Common Crawl, C4, GitHub, Wikipedia, books, and scientific articles. The dataset comprises around 1.4 trillion tokens, carefully curated to prioritize high-quality content and filtered to avoid duplicates and irrelevant information. Parameter sizes in the first version included 7 billion, 13 billion, 33 billion, and 65 billion~\cite{touvron2023llama1}.

The improved LLaMA 2 model, released later, introduced enhanced context sensitivity, dialogue capabilities, and alignment with user preferences. LLaMA 2 scales up to 69 billion parameters and is trained on 2 trillion tokens of publicly available data, allowing it to handle longer, more complex inputs. Key advancements include grouped-query attention (GQA) for larger models, which enhances inference scalability. LLaMA 2 also includes specialized fine-tuning for dialogue through supervised training and Reinforcement Learning with Human Feedback (RLHF). The dialogue-optimized variant, LLaMA 2-Chat, incorporates techniques such as red-teaming, safety tuning, and rejection sampling to improve alignment with human expectations for helpfulness and safety~\cite{touvron2023llama2}.

In 2024, Meta introduced LLaMA 3, representing a new generation of foundation models designed to support multilingual capabilities, advanced reasoning, tool use, and multimodal functionality across text, image, and speech. With up to 70.6 billion parameters, LLaMA 3 was trained on a 15.6 trillion token dataset. Compared to earlier versions, LLaMA 3 features an improved data curation pipeline for pre-training and post-training, enhancing its language understanding and complex reasoning skills. Subsequent updates, including LLaMA 3.1 and LLaMA 3.2, have added further features and fine-tuned performance. For example, LLaMA 3.2, which reaches a maximum of 405 billion parameters, shows strong capabilities in handling long-context applications and demonstrates reduced rates of hallucination.

Similar to other large language models, LLaMA models offer a range of opportunities when integrated with smart environments, particularly for handling complex tasks. For example, the study in~\cite{yin2024harmony} introduces \enquote{Harmony}, an intelligent home assistant system powered by the LLaMA 3-8B model, designed to maintain user privacy and operate locally without requiring an Internet connection. Harmony's architecture consists of three components: a Message Handler, an Agent, and a Controller. The Message Handler processes sensor data and user commands, inferring user needs through both short-term and long-term memory functions. The Agent then formulates action plans based on these inferences, consulting memory for contextual relevance. Finally, the Controller translates the Agent's plans into JSON-formatted commands to control devices, ensuring actions align with the smart home's setup. Harmony demonstrates high accuracy (about 90\%) in executing tasks, comparable to cloud-based solutions such as GPT-4, while significantly reducing hallucination rates. Harmony exemplifies how small-scale LLaMA models can enable effective, privacy-preserving applications for smart spaces without even the need for cloud resources.

\subsection{Google's Bidirectional Encoder Representations from Transformers (BERT)} 
The BERT is another significant example of a large language model (LLM) pre-trained on an extensive text dataset, including sources such as Wikipedia and BookCorpus. The larger version, BERT-Large, contains up to 340 million parameters, enabling it to understand and process language with advanced bidirectional context analysis, where words are analyzed in relation to both their left and right contexts within a sentence~\cite{devlin2018bert}. This is achieved through the transformer architecture, which uses self-attention layers to capture complex relationships between words. 
The study in~\cite{akbari2022transformer} demonstrates the use of BERT to detect behavioral changes in elderly adults living in smart homes, improving monitoring and personalized care. By utilizing BERT’s sequence modeling capabilities, the study identifies anomalies that may indicate health or behavioral changes. To do this, BERT processes sequences of activities as tokenized events (e.g., \enquote{sleep} or \enquote{meal preparation}) and uses its bidirectional attention layers to model typical behavior patterns, enabling it to flag deviations that might indicate health deterioration. 
Another study in~\cite{sun2021bert} employs BERT to enhance indoor positioning accuracy in smart spaces and defend against adversarial attacks. This research presents a crowdsourced indoor localization system that utilizes Bluetooth Low Energy (BLE) fingerprints. BERT’s self-attention mechanism captures complex patterns within BLE signal data, distinguishing authentic fingerprints from adversarial ones. The result is high localization accuracy and improved resilience against database and online attacks, supporting secure and reliable indoor localization.

\textbf{The Gemini family}, developed by Google, represents a powerful series of multimodal LLMs that excel in understanding and processing image, audio, video, and text data. Gemini models are available in various sizes—Ultra, Pro, and Nano—catering to a range of applications from complex reasoning tasks to memory-constrained, on-device use cases. Developers can leverage Gemini's capabilities through the Vertex AI Gemini API and Google AI Gemini API, enabling seamless integration into diverse applications~\cite{team2023gemini, team2024gemini}. 
In the literature, for instance, Gemini has been employed to optimize HVAC control in smart office environments. By integrating real-time environmental data, including temperature, illuminance, and occupant location from office sensors, Gemini's multimodal capabilities support enhanced energy efficiency and occupant comfort. The system processes these multimodal inputs to dynamically predict optimal HVAC setpoints, balancing energy savings with thermal comfort. Results from this experiment demonstrate that the Gemini-based system achieved up to a 47.92\% reduction in power consumption and a 26.36\% improvement in occupant comfort compared to traditional control methods. These findings underscore Gemini's effectiveness in managing complex, real-world environments, illustrating its potential for applications like energy-efficient management in smart spaces~\cite{sawada2024office}.

\subsection{Other LLM Models}
In addition to the most widely recognized LLM models discussed earlier, an increasing number of new models are being introduced, each with specific features and advanced functionalities. Integrating these models within smart environments can offer advantages such as detecting sensor anomalies, enhancing user privacy, and reducing processing power demands.

One example is Claude, a family LLMs developed by Anthropic, an AI research company focused on creating safe and reliable AI. Claude can be used to build AI applications that support various functions, such as engaging in conversations, brainstorming ideas, and analyzing documents. Notably, Claude has demonstrated strengths in handling sensitive topics and maintaining consistency across extended conversational threads. In therapeutic settings, such as ADHD (Attention Deficit Hyperactivity Disorder) support, Claude has been employed as a virtual therapy assistant, fostering an environment that validates patients' emotions and experiences~\cite{berrezueta2024exploring}. This example illustrates the potential of integrating Claude models with indoor robots to assist with a range of applications in smart environments, beneficial in enhancing user comfort and engagement. 

Another example is ChatGLM, developed by Zhipu AI, a customized language model designed for AI-driven applications. ChatGLM is pre-trained using an autoregressive blank-filling objective and can be fine-tuned for various natural language understanding and generation tasks. For instance, the study in~\cite{zhong2024casit} showcases ChatGLM in a multi-agent AI system, where each AI agent corresponds to an LLM model, enabling intelligent processing of complex IoT data in a collaborative manner. This system assigns specific roles to AI agents—such as data analysis and decision-making—allowing them to handle diverse data inputs, including temperature, humidity, and image data. Additionally, the system integrates functions for memory management, summarization, and classification to streamline communication between agents. As a result, the multi-agent LLM deployment system outperforms single-agent configurations in IoT environments, optimizing data processing and minimizing errors. This approach enhances accuracy and reliability for real-time monitoring and anomaly detection across distributed sensor networks.

\subsection{Small and Tiny Language Models}
These models often referred to SLMs or on-device LLMs, are gaining attention for their ability to shift processing tasks directly onto devices, reducing dependence on the cloud and improving latency, data localization, and personalized user experiences. These on-device models, typically containing fewer parameters (e.g., around 10 billion), are optimized for edge devices and support the development of responsive technologies like smart environments. Key examples of these models include Gemini Nano, Nexa AI Octopus, Apple OpenELM, Ferret-v2, Microsoft Phi, MiniCPM, Gemma, LLaMA, ChatGLM, Qwen, Yi, Mistral, and InternLM~\cite{xu2024device}. 

In the literature, models like Gemma (2B) and Phi-2 (2.7B) have been employed to create human-centric smart devices with a focus on privacy and user-friendly interactions. These models allow devices to respond to loosely defined commands and explain actions independently of cloud connectivity. Their operations follow a five-step process, including state modeling, synthetic data generation, and fine-tuning, enabling small LLMs to interpret and respond effectively to user commands on-device, even on compact hardware such as a Raspberry Pi. Case studies on devices such as lamps and thermostats illustrate how these models efficiently adapt to varied user requests, providing meaningful responses and actions based on embedded knowledge, which demonstrates the potential for SLMs in smart environments while bypassing the computational demands of larger models~\cite{king2024thoughtful}.

MobileLLM is another on-device language model optimized for AI applications specifically designed for mobile devices. MobileLLMs leverage advanced techniques such as deep-thin architecture, layer sharing, and grouped query attention. The deep-thin architecture emphasizes model depth (the number of layers) over width (layer dimensions) in sub-billion parameter models, enhancing performance on zero-shot commonsense reasoning tasks~\cite{liu2024mobilellm}. In smart spaces, implementing MobileLLMs on widely used and versatile mobile devices would enable seamless device-to-environment interactions and provide an accessible medium for user engagement within the space.

In addition to SLMs, tiny language models are designed with even fewer parameters—often starting at around one billion—to operate in resource-constrained environments where both computational power and memory are limited. Examples of tiny models include TinyBERT~\cite{jiao2019tinybert}, TinyLlama-1.1B~\cite{zhang2024tinyllama}, MobileLlama-1.4B, Qwen-1.8B, PanGu-1B, and Phi-2.7B~\cite{tang2024rethinking}. These tiny models typically use methods such as model pruning, quantization, knowledge distillation, and efficient architecture design. Model pruning reduces the model size by eliminating unnecessary weights, while quantization represents weights and activations with lower data types (e.g., using 8-bit integers instead of 32-bit floating points). Knowledge distillation trains smaller models to emulate larger ones, and efficient architecture design creates models requiring fewer computations and less memory. By incorporating these techniques, small and tiny language models deliver improved performance on resource-constrained devices, reducing energy consumption and broadening AI adoption. In smart spaces, tiny language models can be implemented on resource-constrained, battery-powered sensors and IoT devices which have limited processing capabilities.

\subsection{Challenges and Limitations}

\textbf{Privacy Risks}: LLMs have shown a tendency to unintentionally disclose personally identifiable information (PII) from both their training data and user inputs, posing significant privacy risks. For instance, LLMs can replicate PII directly from user inputs, meaning that sensitive information—such as personal or health data—may inadvertently appear in generated outputs, even when privacy compliance is emphasized~\cite{priyanshu2023chatbots}. Models like ChatGPT, for example, may retain and reveal names, addresses, or other private details, despite efforts to filter this information. 
This issue is particularly concerning in smart spaces, where LLMs interact dynamically with user data, amplifying the potential for privacy breaches if PII is not well-managed. Although existing privacy-preserving methods aim to reduce these risks, they have limitations~\cite{mireshghallah2024trust}.  Differential privacy algorithms, for example, propose adding noise into the data to minimize memorization of specific data points and reduce PII retention. In addition, Hybrid approaches, which combine LLMs with structured privacy-preserving modules, employ techniques such as PII detection, redaction, and differential privacy at the inference stage. However, these methods often create a trade-off between privacy and model performance, adding computational overhead that may not be feasible for the low-latency requirements of smart spaces. 
Ultimately, privacy concerns in LLMs highlight the need for innovation in model architectures. Developing solutions that protect sensitive data while maintaining efficiency and compliance will be essential for advancing LLM applications in sensitive environments such as smart spaces.

\textbf{Security Risks:} Malicious attacks may exploit LLMs to bypass detection mechanisms, manipulate outputs, or compromise the integrity of smart environments. One of the major threats is adversarial attacks, where attackers subtly modify input data to influence the model’s responses. Even small alterations, such as minor word substitutions or paraphrasing, can bypass common monitoring approaches such as classifiers and watermarking systems that are typically used to secure LLM outputs. Attackers may also craft prompts that instruct LLMs to generate responses in a style or vocabulary that evades detection~\cite{shi2024red}. 
To mitigate these vulnerabilities, methods such as instructional prompt filtering, which adjusts input processing to detect adversarial patterns, and advanced watermarking, which embeds distinctive markers in LLM outputs, have shown promise in enhancing model robustness. However, these approaches add computational overhead and are not entirely foolproof. This highlights the need for lightweight, scalable security solutions that protect LLMs while meeting the real-time needs of smart environments.

\textbf{Limitations in Reasoning and Planning:} Despite their proficiency in generating human-like text, LLMs exhibit critical limitations in reasoning and planning capabilities, which are essential for real-world applications like smart spaces. In these environments, consistent and reliable decision-making based on real-time data is paramount, yet LLMs often produce unpredictable or inaccurate outputs due to their reliance on patterns from training data rather than true logical inference or causal understanding. LLMs struggle to generalize reasoning processes across varying contexts, often resulting in flawed or overly simplistic responses. Unlike symbolic systems that follow explicit logical rules, LLMs rely on predicting token sequences based on statistical patterns, limiting their ability to engage in the conditional or iterative reasoning necessary for complex, multi-step deductions. For instance, models like BERT focus on superficial patterns rather than learning transferable logical rules, leading to reasoning errors when encountering unfamiliar data~\cite{zhang2022paradox}. 
In addition to reasoning limitations, LLMs face significant challenges in planning, which is crucial for dynamically adapting in smart spaces. While LLMs can mimic structured reasoning in sequence generation, they often lack an understanding of causality or state-dependent conditions. This leads to failures in generating executable plans in real-world settings~\cite{valmeekam2022large}. Models such as GPT-4 and Claude demonstrate limited success in producing consistent and valid plans autonomously, particularly in scenarios that deviate from their training examples. Their inability to self-verify or correct plans further restricts their application in tasks requiring high precision and reliability~\cite{valmeekam2024llms}.

To address these limitations, neurosymbolic and hybrid frameworks have been proposed. For instance, LLM-Modulo frameworks integrate symbolic reasoning tools to improve both reasoning and planning robustness. However, these methods add computational overhead and complexity, posing practical barriers for real-time applications in smart spaces, where efficiency and responsiveness are essential~\cite{borazjanizadeh2024reliable}. 

Despite advancements in neurosymbolic frameworks, they face challenges with tasks requiring multi-step logic, showing significant performance drops as complexity increases. Their inability to filter out irrelevant information amplifies errors when extraneous context is introduced~\cite{mirzadeh2024gsm}. These limitations address the need for hybrid frameworks that combine traditional AI techniques with the capabilities of LLMs, incorporating structured reasoning components to address the demands of smart environments ~\cite{kambhampati2024llms}.

Fortunately, the advancements in Large Reasoning Models (LRMs), such as OpenAI's o1 model, can overcome these reasoning limitations of LLMs. By leveraging techniques like Chain-of-Thought (CoT) fine-tuning and reinforcement learning, the o1 model demonstrates enhanced proficiency in solving complex reasoning tasks, including multi-step logic problems and mathematical computations. For example, the Marco-o1 LRM~\cite{zhao2024marco} has been successfully applied to real-world problem-solving tasks, showcasing its potential for tackling complex challenges. This capability aligns with the needs of smart spaces, where dynamic and adaptive decision-making is critical.

\textbf{Hallucination Risk:} Hallucinations in LLMs refer to instances where the model generates information that is not based in factual or accurate data. This phenomenon occurs when the model produces content absent from the input data, often influenced by internal biases or overconfidence. Hallucinations pose significant risks in applications requiring precise and contextually accurate information~\cite{zhang2023siren}. 
Hallucinations persist across LLM generations due to the probabilistic nature of token prediction and the lack of true comprehension. Even with extensive training data, LLMs show limitations in fact-verification and struggle to differentiate between high-confidence predictions and unverified associations~\cite{xu2024hallucination}. Techniques such as retrieval-augmented generation (RAG), which allows LLMs real-time access to authoritative databases, and real-time verification frameworks such as EVER, which detect and correct inaccuracies as they arise, have shown promise in reducing hallucination rates. However, these methods increase computational demands, which can hinder their practicality in latency-sensitive applications~\cite{tonmoy2024comprehensive}.  
In smart spaces, where ongoing and accurate interaction with the physical world is essential, hallucinations can introduce substantial risks, pinpointing the need for solutions that ensure LLMs provide reliably accurate outputs in real-time contexts.

\textbf{Computational Overhead, Real-Time Constraints, and Energy Demands:} The use LLMs in smart environments places significant demands on processing power, memory, and response time. These requirements pose particular challenges for resource-constrained edge devices, which often lack the capacity to meet the high computational needs of LLMs, leading to increased latency~\cite{li2023transformer}. To address these limitations, strategies such as model compression, edge-cloud collaboration, the use of smaller and tiny language models, and quantization techniques are under development to better manage computational resources.

The latency associated with generating outputs from LLMs can compromise their effectiveness in smart spaces. To mitigate this, hybrid approaches have been proposed that integrate LLMs with locally deployed sensors to perform preliminary inference, before transferring data to cloud-based LLMs for more complex analysis. This approach would reduce latency and enhance responsiveness~\cite{sun2024ai}. However, these approaches are still in early stages and may not be suitable for some smart space applications that require processing large datasets.

LLMs also have high energy demands, consuming substantial power during both training and inference. For example, studies have shown that models such as LLaMA require significant computing and memory resources across multiple GPUs, resulting in high energy consumption and increased carbon emissions~\cite{samsi2023words}. In response, studies introduce carbon-efficient architectures, such as distributed computing and parallel execution, to help reduce computational loads. However, these architectures present practical challenges in real-time applications for smart spaces~\cite{faiz2023llmcarbon}. Distributed models require extensive inter-device communication, which increases network traffic and bandwidth utilization. Hyperparameter tuning has shown promise as an alternative, reducing unnecessary computations by optimizing model parameters to lower power consumption during training and inference. However, tuning itself can require significant computation, which may not align well with the energy constraints of certain smart space applications~\cite{strubell2020energy}. In summary, while LLMs hold promise for enhancing smart spaces, their deployment is constrained by energy-related challenges that may limit their practical use in these settings.

\section{Concluding Remarks}
%=====================
\label{sec:Conclusion}

\subsection{Remarks}

The development of AI-driven smart spaces enables human-like decision-making by dynamically optimizing energy use, comfort, and security. However, achieving such environments presents a range of challenges, as these spaces must proactively adapt to users' needs. These key challenges include managing diverse sensor and network devices, implementing effective communication protocols, handling extensive datasets, developing efficient AI/ML models to interpret data and adapt to user behavior, ensuring ongoing sensor calibration and maintenance, and upholding robust data privacy and security mechanisms.

Creating effective smart spaces requires the careful selection of suitable sensor devices, communication gateways, and computing platforms tailored to the specific demands of these environments. A diverse range of sensor technologies is available, each offering different levels of accuracy and requiring distinct deployment strategies to maintain reliable data communication. Ensuring high-quality data collection (via regular sensor calibration) and maintaining robust connectivity across sensor networks are critical for uninterrupted data flow. Since users are central to smart space functionality, secure privacy measures, such as encryption and access control, are essential to protect sensitive user information. 
In addition, selecting suitable computing methodologies, whether edge-based or cloud-based, is crucial for balancing processing, communication, and power demands. Depending on data size and application requirements, data can be processed locally on edge servers to deliver real-time services, thereby reducing the need for extensive data transmission to the cloud and alleviating privacy concerns.

AI plays a pivotal role in creating interactive environments within smart spaces, where the system anticipates and responds to users' needs. Implementing efficient AI/ML methodologies is vital, as current ML and DL models are well-suited to tasks such as activity recognition, occupancy, and anomaly detection. These models can handle tasks within the constraints of processing, communication, and power resources available in most smart environments. 
For advanced AI models, such as transformer networks and LLMs, energy consumption presents a significant challenge, as these models require large datasets and extensive computational power. However, hybrid processing approaches, which involve deploying smaller, task-specific AI models at the edge and reserving complex computations for cloud-based models, offer a promising solution. This strategy balances responsiveness with computational demands, thereby enhancing real-time capabilities for smart spaces.

The LLMs as emerging techniques offer significant advantages, making them potential candidates for integration in smart spaces due to their multimodal capabilities and scalability. Their ability to process and integrate diverse data types—such as text, audio, and visual inputs—enables more intuitive and context-aware user interactions. For example, LLMs can interpret natural language commands, enabling intuitive user interactions and converting raw sensor data into actionable insights. When integrated with RAG mechanisms, LLMs become even more powerful in supporting decision-making and automation. Their scalability allows for flexible deployment across various devices, facilitating seamless integration within existing smart space. However, despite their rapid advancements, LLMs still face a range of challenges including implementation complexity, high computational demands, energy consumption, reasoning limitations, and real-time processing requirements. 
In addition to LLMs, smaller language models offer practical solutions for basic smart space applications, especially in resource-limited settings or environments with limited connectivity.

In conclusion, ongoing advancements in AI, along with innovations in sensing, communication, and computing technologies, promise the creation of interactive AI-based smart spaces that enhance user experience and promote sustainable living environments.

\subsection{Future Directions}

To further enhance smart spaces, future research should address the interoperability of heterogeneous IoT devices by developing standard protocols and interfaces, as well as resource allocation mechanisms that facilitate seamless communication across varied devices and networks. Additionally, efforts should focus on developing efficient algorithms, particularly through model compression and hardware-aware designs, to optimize AI models for resource-constrained IoT devices.

Increasing the adaptability of AI models to different indoor environments —such as homes, offices, and healthcare facilities— will broaden their applicability. Future research should explore cross-domain adaptation techniques that function efficiently across diverse settings, minimizing the need for environment-specific configurations.

The development of efficient digital twin tools capable of integrating AI/ML and LLMs into smart spaces represents a significant research direction. Such a digital twin would enable real-time simulations and automated decision-making within AI-powered smart environments. Current research often presents static digital solutions, lacking the integration of advanced functional AI and LLM methodologies. This highlights the need for dynamic and adaptive digital twining solutions for AI-powered smart spaces.

Most importantly, as IoT devices in smart spaces become more prevalent and increasingly collect real-time sensitive data, safeguarding user privacy remains a top priority. Developing privacy-preserving methodologies and frameworks for all components creating the smart spaces will be essential to protect user information while enabling advanced AI-powered applications and automation.

% Comment these if \bibliographystyle{unsrt}  is enabled.
\bibliographystyle{IEEEtran}
\bibliography{bibliography}

%\bibliographystyle{unsrt}  

%\begin{thebibliography}{1}

%\bibitem{eriksen2014plastic}
%Marcus Eriksen and Laurent CM. Lebreton and Henry S. Carson  and Martin Thiel and Charles J. Moore and Jose C. Borerro and Francois Galgani and Peter G. Ryan and Julia Reisser.
%\newblock Plastic pollution in the world's oceans: more than 5 trillion plastic pieces weighing over 250,000 tons afloat at sea.
%\newblock In {\em PloS one}, vol. 9, no. 12, p. e111913, 2014.

%\end{thebibliography}

\end{document}